\pdfoutput=1

\documentclass[11pt]{article}

\usepackage[preprint]{acl}

\usepackage{times}
\usepackage{latexsym}
\usepackage{mathrsfs}
\usepackage{amsmath}
\usepackage{amssymb}
\usepackage{bbm}
\usepackage{booktabs}
\usepackage{graphicx}
\usepackage{lscape}
\usepackage{hyperref}

\usepackage{algorithm}
\usepackage[noEnd=false]{algpseudocodex}
\usepackage{graphicx}
\usepackage{caption}
\usepackage{subcaption}
\usepackage{multirow}
\usepackage{makecell}

\usepackage{listings}

\usepackage[T1]{fontenc}

\usepackage[utf8]{inputenc}

\usepackage{microtype}

\usepackage{inconsolata}

\usepackage{graphicx}

\newcommand\benchmarkname{\textcolor{black}{\textsc{TextGames Benchmark}}}

\newcommand\benchmarknameonly{\textcolor{black}{\textsc{TextGames}}}

\algrenewcommand\algorithmicensure{\textbf{Initialize:}}

\title{\benchmarknameonly{}: Learning to Self-Play Text-Based Puzzle Games\\via Language Model Reasoning}

\author{
  Frederikus Hudi\thanks{\footnotesize{Equal contributions.} $^\dagger$\footnotesize{The work was done outside Capital~One.}}$^{1}$, Genta Indra Winata$^{*2,\dagger}$, Ruochen Zhang$^{*3}$, Alham Fikri Aji$^{*4}$
  \\
  $^{1}$NAIST  \quad
  $^{2}$Capital One  \quad
  $^{3}$Brown University  \quad
  $^{4}$MBZUAI  \quad
  \\
  \texttt{frederikus.hudi.fe7@naist.ac.jp}  \quad
  \texttt{genta.winata@capitalone.com}  \\
  \texttt{ruochen\_zhang@brown.edu}  \quad
  \texttt{alham.fikri@mbzuai.ac.ae}  \\
  }

\begin{document}

\maketitle
\begin{abstract}
Reasoning is a fundamental capability of large language models (LLMs), enabling them to comprehend, analyze, and solve complex problems. In this paper, we introduce $\benchmarknameonly$, an innovative benchmark specifically crafted to assess LLMs through demanding text-based games that require advanced skills in pattern recognition, spatial awareness, arithmetic, and logical reasoning. Our analysis probes LLMs' performance in both single-turn and multi-turn reasoning, and their abilities in leveraging feedback to correct subsequent answers through self-reflection. Our findings reveal that, although LLMs exhibit proficiency in addressing most easy and medium-level problems, they face significant challenges with more difficult tasks. In contrast, humans are capable of solving all tasks when given sufficient time. Moreover, we observe that LLMs show improved performance in multi-turn predictions through self-reflection, yet they still struggle with sequencing, counting, and following complex rules consistently. Additionally, models optimized for reasoning outperform pre-trained LLMs that prioritize instruction following, highlighting the crucial role of reasoning skills in addressing highly complex problems. 
\end{abstract}

\begin{figure}[!th]
    \centering
    \includegraphics[width=\linewidth]{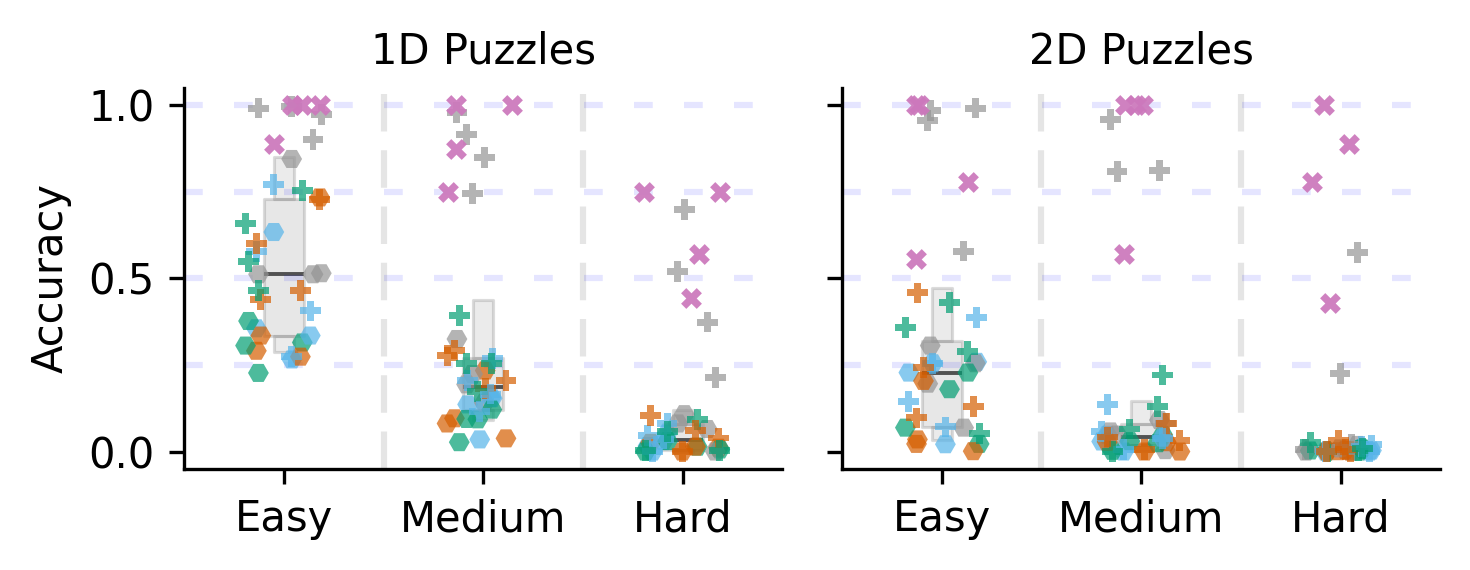}
    \includegraphics[width=\linewidth]{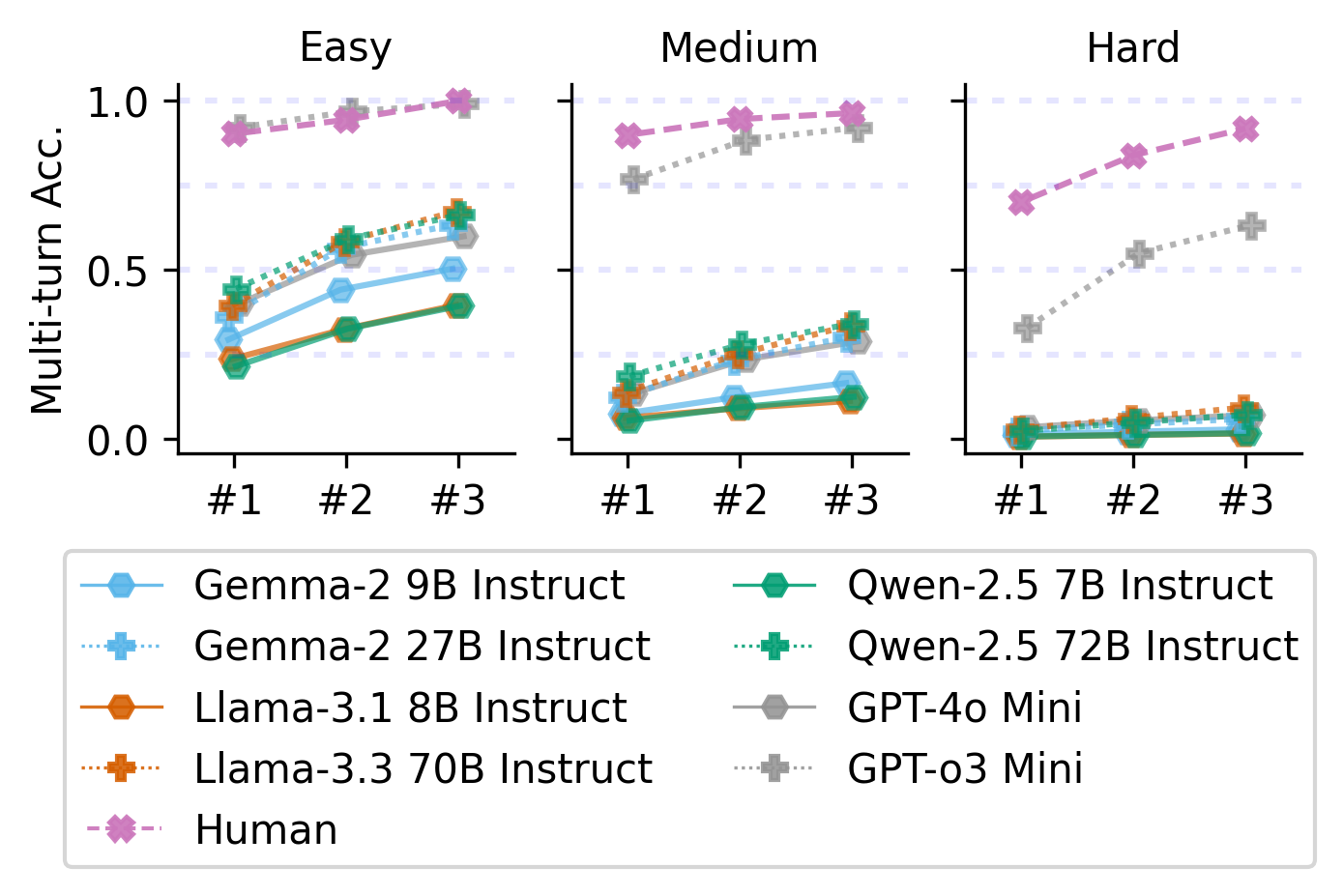}
    \caption{Single-turn performance on $\benchmarknameonly$ games across 1D and 2D Puzzles challenges with varying difficulty levels \textbf{(top)}, alongside the improvement in accuracy achieved through increased turn attempts via self-reflection, with the x-axis representing the number of turns \textbf{(bottom)}.}
    \label{fig:2d_and_level}
\end{figure}

\section{Introduction}
Reasoning is a fundamental skill essential for logical thinking and development, enabling large language models (LLMs) to tackle complex problems~\cite{wei2022chain,longpre2023flan,srivastava2023beyond}. This skill emphasizes the need for creating LLMs capable of handling tasks such as mathematical~\cite{hendrycks2measuring,shao2024deepseekmath,trinh2024solving}, commonsense~\cite{talmor2019commonsenseqa,geva2021did,brohan2023can}, and symbolic reasoning~\cite{nye2021improving,sprague2024cot}. In general, reasoning is a multifaceted ability that involves understanding the context and effectively applying inference to solve problems. Research on LLMs has examined their reasoning capabilities across various dimensions, including their capacity to follow instructions for multi-hop reasoning~\cite{yang2024large}, comprehend psychological concepts~\cite{almeida2024exploring}, and use context in classification tasks~\cite{winata2024miners}, and constrained logical tasks~\cite{zhou2023instruction}. LLMs have also demonstrated remarkable skills in game reasoning, such as solving crossword puzzles~\cite{berruti2024automatic,saha2024language,zugarini2024clue}, physics-based puzzle games~\cite{oh2024langbirds}, and turn-based games~\cite{feng2024chessgpt,guo2024can}.

\begin{figure*}[!th]
    \centering
    \includegraphics[width=\linewidth]{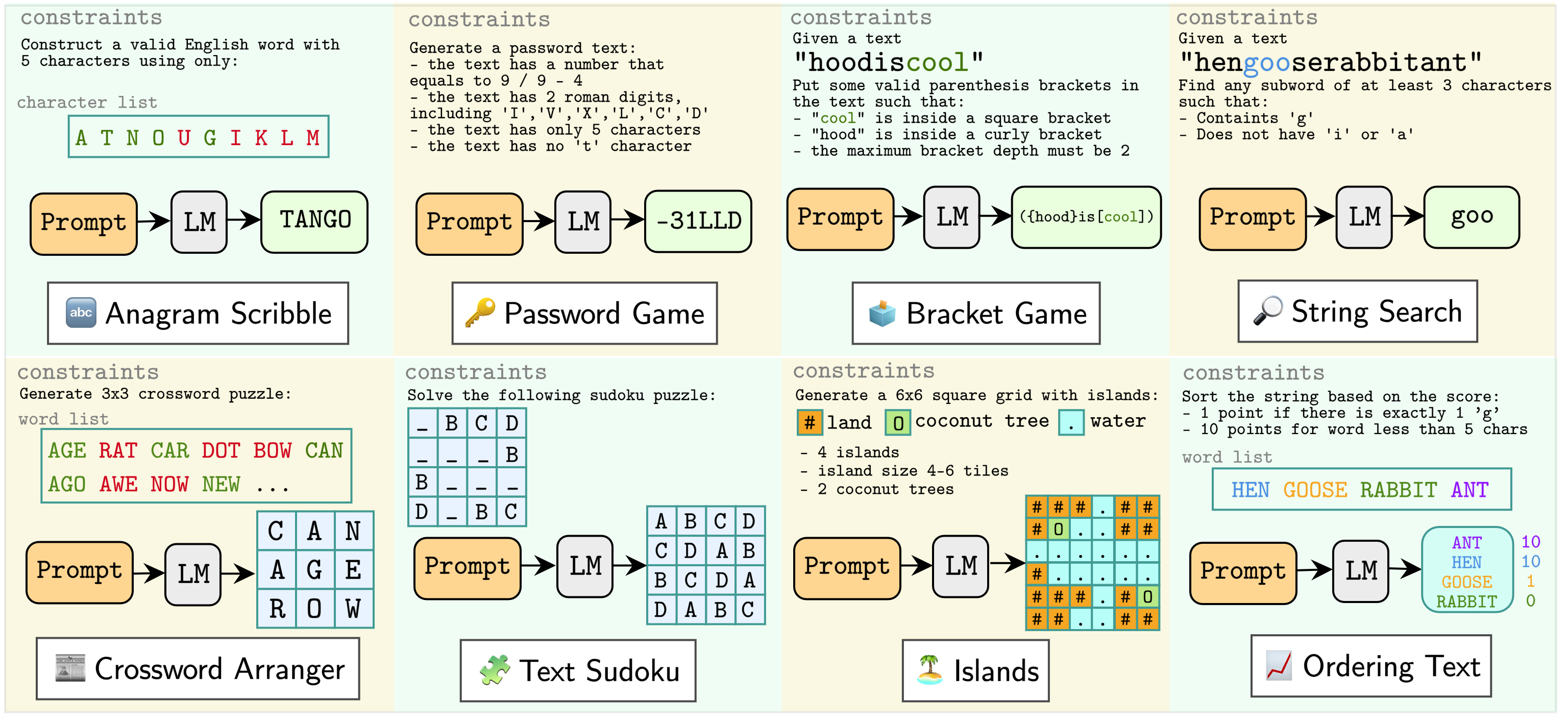}
    \caption{\benchmarkname{} consists of eight text-based puzzle games, each with unique constraints and gameplay mechanics. The top four games are 1D Puzzles, while the bottom four are 2D Puzzles.}
    \label{fig:game}
\end{figure*}

A longstanding issue in reasoning with LLMs is their tendency to hallucinate and inconsistency during inference~\cite{maynez2020faithfulness,ji2023survey,huang2024survey}. Recently, self-reflection techniques have been employed to mitigate these hallucinations and improve the performance of LLMs through multiple rounds of follow-up interactions~\cite{ji2023towards}. Additionally, self-evaluation has been applied to question-answering tasks~\cite{ren2023self}, offering feedback that enables models to correct themselves. Consequently, LLMs have demonstrated the ability to rectify errors across various domains, gradually producing correct answers over successive iterations~\cite{shinn2024reflexion}. Despite these advancements, we aim to further challenge LLMs by engaging them with puzzles that require a combination of skills, including pattern recognition, spatial awareness, arithmetic, and logical thinking.

In our work, we introduce $\benchmarknameonly$, a new benchmark designed to assess the proficiency of LLMs in solving text-based logical puzzle games and performing complex, constraint-based reasoning. The intricate rules of these puzzles allow us to evaluate the LLMs' capacity to follow detailed instructions. Additionally, we investigate whether LLMs can self-reflect on their previous generations when given feedback, correcting their errors by responding to specific error messages and refining their outputs. We also provide performance comparison between reasoning-specialized LLMs, with models that emphasize instruction-following. Our analysis indicates that even the recent advanced LLMs, such as the Llama 70B~\cite{dubey2024llama} and Qwen2 72B Instruct~\cite{yang2024qwen2} models, perform adequately on Easy and Medium levels but struggle at the Hard level. In contrast, models specifically optimized for reasoning, like GPT-o3 Mini, exhibit strong performance on these more difficult tasks, as illustrated in Figure~\ref{fig:2d_and_level}. We hypothesize that this disparity arises because $\benchmarknameonly$ demands a high level of reasoning ability to comprehend the rules and apply a combination of reasoning skills to solve the problems that Instruct models may not fully possess.

In summary, our contributions are threefold:
\begin{itemize}
    \item We introduce $\benchmarknameonly{}$,\footnote{The code can be accessed at~\url{https://github.com/fhudi/textgames}.} a text-based game benchmark that assesses LLMs' various logical reasoning skills. The benchmark features eight puzzle games across three difficulty levels. Figure~\ref{fig:game} offers an overview of the game visualizations.
    \item We perform a thorough evaluation across a range of LLMs, including both off-the-shelf and proprietary models, in zero-shot and one-shot scenarios. We additionally compare their performance with that of human participants.
    \item We demonstrate that LLMs improve when given feedback in multi-turn interaction, enabling them to self-reflect on previous generations. Our observation on reasoning-focused models' performance also reveals that there can be diminishing returns on test-time scaling in some difficult games.
\end{itemize}


\section{\benchmarkname}\label{task}
\begin{table*}[!ht]
\centering
\resizebox{\textwidth}{!}{
    \begin{tabular}{llcccc}
        \toprule
        \multicolumn{2}{c}{\textbf{Task}} & \textbf{Output Format} & \textbf{Category} & \textbf{Skill-sets} & \textbf{Reasoning} \\
        
        \midrule
        \multicolumn{6}{l}{\texttt{1D Puzzles}} \\ 
         & \textbf{Anagram Scribble} & Single-line text & English words & Pattern Recognition, Knowledge & Abductive \\
         & \textbf{Password Game} & Single-line text & Numbers \& Characters & Arithmetic, Knowledge & Abductive \\
         & \textbf{Bracket Game} & Single-line text & Coordinates & Counting & Deductive \\
         & \textbf{String Search} & Single-line text & String Matching &  Logical Thinking & Deductive \\
        
        \midrule
        \multicolumn{6}{l}{\texttt{2D Puzzles}} \\ 
         & \textbf{Crossword Arranger} & 2D-Grid & English words  & Pattern Recognition, Spatial Awareness & Deductive \\
         & \textbf{Text Sudoku} & 2D-Grid & Numbers \& Characters & Spatial Awareness & Deductive \\
         & \textbf{Islands} & 2D-Grid & Coordinates \& Geometry & Spatial Awareness & Abductive \\
         & \textbf{Ordering Text} & Multiple words & Strings \& Sorting & Arithmetic, Comparative & Deductive \\
        
        \bottomrule
    \end{tabular}
    
}
\caption{Detailed information on \benchmarknameonly{} puzzle games, encompassing a broad spectrum of output formats, categories, skillsets, and reasoning types.}
\label{tab:textgames-benchmark-detail}
\vspace{-2mm}
\end{table*}
We introduce our benchmark $\benchmarknameonly$, which comprises eight text-based puzzle games, each featuring three distinct levels of difficulty, aimed at evaluating the reasoning abilities of LLMs. These games are meticulously designed to assess a wide array of reasoning skills, encompassing both abductive and deductive reasoning. Additionally, we differentiate various skills through diverse output formats, as described in Table~\ref{tab:textgames-benchmark-detail}.


\subsection{List of Games}
We provide a detailed definition of the games as follows:
\subsubsection{Anagram Scribble}\label{task:anagram}
Given a list of Latin characters, the player's objective is to arrange them into a valid $N$-character English word, without regard to case sensitivity. We explore two scenarios: one where characters can be used multiple times and another where each character can only be used once.


\subsubsection{Password Game}\label{task:password}
Given a set of rules, the player is challenged to construct a sequence of characters that fulfills all specified requirements, similar to creating a password. These rules involve generating text based on character counts, incorporating English alphanumeric characters, distinguishing between uppercase and lowercase letters, and including special characters and Roman numerals. Additionally, we introduce more complex tasks that require commonsense knowledge, such as identifying the capital city or continent of a specified country. Furthermore, we add simple arithmetic constraints, such as ``The text must include a number equal to seven times six.''



\subsubsection{Bracket Game}\label{task:bracket}
Given a concatenation of several English words, the player is tasked with enclosing segments of the text using four different types of parentheses: `[]', `\{\}`, `()', and `<>'. These brackets must be correctly paired where each open bracket must have a corresponding close bracket, and vice versa. Additionally, there are requirements regarding bracket depth that the player must adhere to.

\subsubsection{String Search}\label{task:string}
Given a random sequence of characters mixed with some valid English words, the player is challenged to find a substring—a consecutive sequence of characters—that meets a specified set of rules. These rules dictate conditions such as the length of the substring, required characters, prohibited characters, and whether the resulting substring must be a palindrome.


\subsubsection{Crossword Arranger}\label{task:crossword}
Given a list of English words, each of length $N$, the player is tasked with arranging these words into a crossword puzzle. Without any repetitions, a total of $2N$ words from the list must be placed in either a horizontal or vertical orientation, forming a connected configuration within an $N \times N$ square grid. Blank cells are not used to separate the words.

\subsubsection{Text Sudoku}\label{task:sudoku}
Given a sparsely filled square grid of size $N^2 \times N^2$, the player is tasked with filling the blank cells with numbers such that no identical numbers appear within the same row, column, or $N \times N$ sub-grid. The player must fill only the blank cells, leaving the pre-filled cells unchanged. We utilize grids with $N$ equal to 2 and 3, meaning the numbers range from 1 to 4 and 1 to 9, respectively. Alternatively, these numbers can be substituted with unique characters; for instance, we experiment with using Latin alphabets `A' to `I' in place of numbers 1 to 9.

\begin{table*}
    \centering
    \resizebox{.9\textwidth}{!}{
    \begin{tabular}{cllll}
        \toprule
        \multicolumn{1}{c}{\textbf{Game}} & \multicolumn{1}{c}{\textbf{Easy}} & \multicolumn{1}{c}{\textbf{Medium}} & \multicolumn{1}{c}{\textbf{Hard}}  \\
        
        \midrule
        \multirow[l]{3}{*}{\makecell{Anagram\\Scribble}}
        & - 3 to 5 letter English word & - 6 to 7 letter English word & - 8 to 10 letter English word \\
        & - Character list $\leq$ 10 & - Character list $\leq$ 10 & - Character list $\leq$ 10 \\
        & - Repeatable use of char & - Repeatable use of char & - Each char can only be used once\\
        
        \midrule
        Password
        & - 2 Rules & - 4 Rules & - 6 Rules \\
        
        \midrule
        \multirow[l]{3}{*}{\makecell{Bracket\\Game}}
        & - Rules = 3 & - Rules = 5 & - Rules = 5 \\
        & - Words = 3 & - Words = 5 & - Words = 5 \\
        & - Depth = 2 & - Depth = 2 & - Depth = 3 \\
        
        \midrule
        \multirow[l]{4}{*}{\makecell{String\\Search}}
        & - Text length $\leq$ 10 characters & - Text length $\leq$ 20 characters & - Text length $\leq$ 40 characters \\
        & - At most 2 constraints & - At most 3 constraints & - At most 5 constraints \\
        & - Multiple solutions may exist & - Multiple solutions may exist & - Single solution \\
        & - No complex rules &  - No complex rules & \\ 
        
        \midrule
        \makecell{Crossword\\Arranger}
        & \makecell[l]{- Board size = 3x3 \\- Words =  8 \\- 25\% Noise words}
        & \makecell[l]{- Board size = 4x4 \\- Words = 16 \\- 50\% Noise words}
        & \makecell[l]{- Board size = 5x5 \\- Words = 20 \\- 50\% Noise words}
        \\
        
        \midrule
        \multirow[l]{2}{*}{\makecell{Text\\Sudoku}}
        & - Board size = 4x4 & - Board size = 4x4 & - Board size = 9x9 \\
        & - Empty ratio = 0.25 & - Empty Ratio = 0.5 & - Empty Ratio = 0.4 \\
        
        \midrule
        \multirow[l]{3}{*}{Islands}
        & - Only 1 island & - 1 to 3 islands & - 3 to 6 islands \\
        & - No coconut tree & - No complex constraints & \\
        & - No complex constraints\\
        
        \midrule
        \makecell{Ordering\\Text}
        & \makecell[l]{- Rules = 2 \\- Words = 3 \\- 3 $\leq$ Word Length $\leq$ 8}
        & \makecell[l]{- 2 $\leq$ Rules $\leq$  4\\-  4 $\leq$ Words $\leq$  6\\- 3 $\leq$ Word Length $\leq$  8}
        & \makecell[l]{- 4 $\leq$ Rules $\leq$  8\\-  6 $\leq$ Words $\leq$ 10\\- 3 $\leq$ Word Length $\leq$ 15}
        \\
        
        \bottomrule
    \end{tabular}
    }
    \caption{Difficulty levels of $\benchmarknameonly{}$ puzzle games detailed with associated constraints and rules.}
    \label{tab:difficulty-levels}
    \vspace{-2mm}
\end{table*}

\subsubsection{Islands}\label{task:islands}
Given a grid size of $N$, along with a specified set of rules, the player must construct an $N \times N$ square grid using the characters `\texttt{.}', `\texttt{\#}', or `\texttt{o}', which represent water, land, and coconut trees, respectively. A contiguous group of land tiles connected in the four cardinal directions forms an island. The task requires adherence to all rules, which govern the number of islands, the size of each island, and the allowable number of coconut trees.

\subsubsection{Ordering Text}\label{task:ordering}
Given a set of scoring rules and a list of words, the player is tasked with sorting the list from the highest-scoring word to the lowest. The scoring rules encompass checks for the presence of specific character sequence patterns, the length of the words, as well as the prefixes and suffixes of the words. Points in each scoring rule can range from $-100$ to $100$.


\subsection{Challenges and Difficulty Levels}

For comprehensive details about the games, including formats, categories, and the reasoning skills required, please refer to Table~\ref{tab:textgames-benchmark-detail}. Each game is designed with three levels of difficulty: Easy, Medium, and Hard, with specifics available in Table~\ref{tab:difficulty-levels}. The difficulty escalates through factors like the increased size of a 2D board, more stringent constraints, and progressively challenging reasoning tasks. Most games are designed to support multiple solutions, which can vary with the difficulty level. For instance, in Anagram Scribble (\ref{task:anagram}), the same set of characters can be rearranged to create different English words, such as ``game'' and ``mega.'' In Islands (\ref{task:islands}), the location of coconut trees can be arbitrary. In contrast, Ordering Text (\ref{task:ordering}) offers only one possible solution, as words with the same score are sorted lexicographically.

\subsection{Game Categories}

The benchmark tasks can be divided into two categories: 1D and 2D formats. The 1D puzzles include Anagram Scribble (see Section \ref{task:anagram}), Password Game (see Section \ref{task:password}), Bracket Game (see Section \ref{task:bracket}), and String Search (see Section \ref{task:string}). In contrast, the 2D puzzles demand spatial awareness and the capacity to track values across multiple rows. These include Crossword Arranger (see Section \ref{task:crossword}), Text Sudoku (see Section \ref{task:sudoku}), Islands (see Section \ref{task:islands}), and Ordering Text (see Section \ref{task:ordering}). Generally, models demonstrate superior performance on 1D puzzles. For instance, the performance of LLMs on easy 2D puzzles is comparable to their performance on medium-difficulty 1D puzzles, while their performance on medium 2D puzzles parallels that on hard 1D puzzles. This is illustrated in Figure \ref{fig:2d_and_level}, highlighting the challenges LLMs face with 2D spatial reasoning.

\subsection{Game Generation}
For each game, we create instances by randomly sampling according to the specified rules for each difficulty level, resulting in 1,000 test samples per difficulty. This amounts to a total of 24,000 test samples across all games and difficulty levels. Additionally, we generate a number of training samples for few-shot learning across all difficulties, ensuring that these samples do not overlap with the test set. We refer to the test samples as $\mathscr{D}^\text{test}$ and the training samples as $\mathscr{D}^\text{train}$.

\section{$\benchmarknameonly$ Evaluation}

For our $\benchmarknameonly$, we design a game evaluation framework where LLMs emulate player behavior to play the games. This system uses a LLM to generate solutions and integrates a grader to verify their correctness. To further test models' performance, we implement multi-turn prompting, enabling the model to iteratively refine its responses. This iterative process involves receiving feedback from the grader, which allows the models to self reflect and attempt to correct the answers.

\subsection{Prompt Generation}
We utilize in-context learning prompts to generate answers and evaluate the capabilities of LLMs under two configurations: zero-shot and one-shot prompts. Our prompt is defined as $P \gets (T, C, E, I)$, where it is constructed using a prompt template $T$, along with constraints $C$, one-shot examples $E$, and relevant context $I$ from previous interactions for multi-turn scenarios. We denote the LLMs used for inference as $\theta$ and the grader that evaluates the correctness of the answers as $\mathcal{G}$. Detailed information about the prompts for each game is provided in Appendix~\ref{sec:task_details}.

\subsection{Multi-Turn Prompting}


Algorithm~\ref{alg1} outlines the procedure for implementing multi-turn prompting, a strategy that iteratively refines responses based on feedback from a grader. At each turn, the model generates a response given the test constraint, few-shot examples, and previous interactions. The grader evaluates the response and provides feedback if errors are detected. The interaction history is updated with both the response and feedback, allowing the model to adjust its outputs in subsequent turns. The process terminates early if the grader confirms a correct response, ensuring adaptability while enabling iterative refinement. A complete list of feedback for all games can be found in Appendix~\ref{sec:feedbacks}.

\begin{figure*}[!th]
    \centering\includegraphics[width=\textwidth]{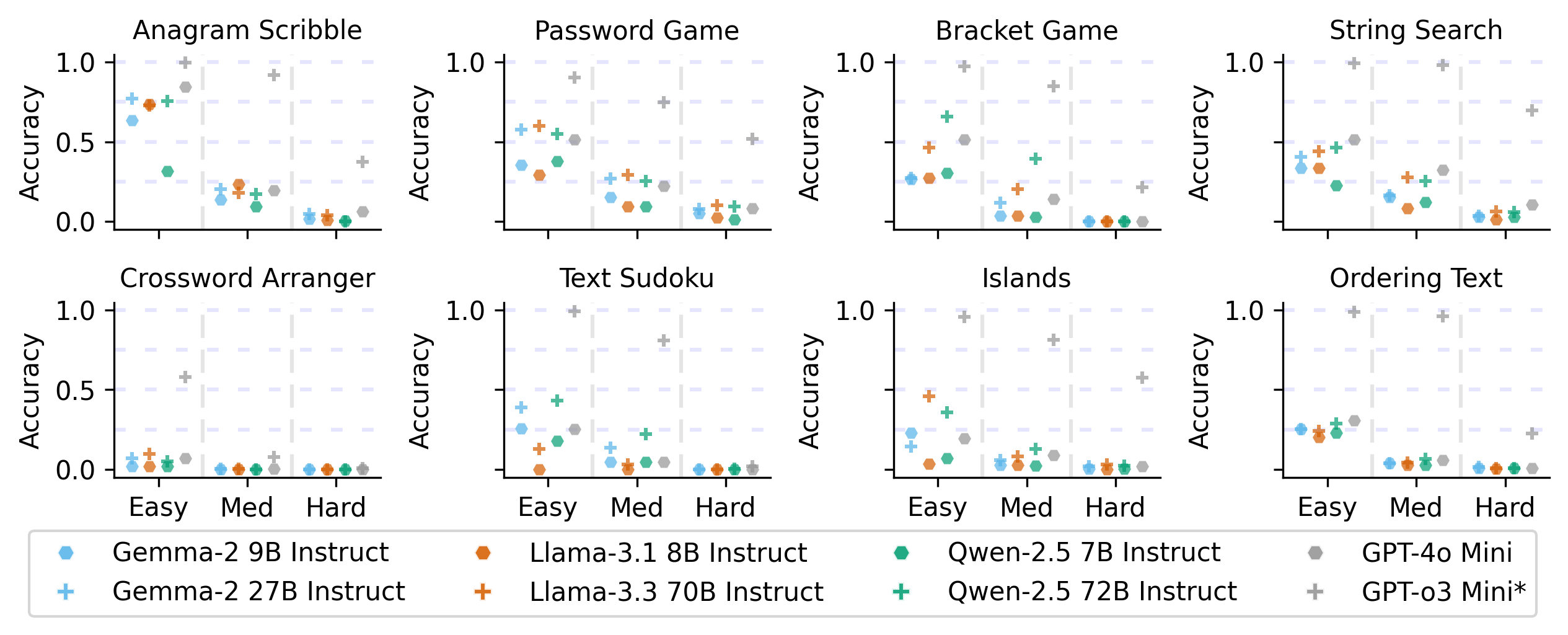}
    \caption{
    LLM Results on $\benchmarkname$ in the one-shot setting. Med indicates Medium-difficulty level. \text{*}For GPT-o3 Mini, we present the results from zero-shot setting.
    }\label{results:1_shot}
    \vspace{-3mm}
\end{figure*}

\section{Experimental Setup}

For each task described in Section~\ref{task}, we begin by developing a grader to verify the correctness of the answers. These graders function similarly to those used on online judge platforms or in competitive programming contests, focusing solely on determining whether an answer is correct or incorrect. Subsequently, we evaluate the performance of various LLMs using these graders. Additionally, we have created a web-based platform to collect data for testing human performance on the same tasks, allowing for a comprehensive comparison between human and model capabilities.

\begin{algorithm}[!t]
\small
\caption{$\benchmarknameonly$ Evaluation System}
\label{alg1}
\begin{algorithmic}[1]
\Require LLM $\theta$, Grader $\mathcal{G}$, Template $\mathcal{T}$, Dataset $\mathscr{D}^{\{\text{train},\text{test}\}}$.
\Ensure Few-shot example(s) $E \subseteq \mathscr{D}^{\text{train}}$.
\Ensure Maximum Turn $N = 3$.
\ForAll{Constraints $C \in \mathscr{D}^{\text{test}}$}
    \State $I \gets [\ ]$
    \For{$i = 1, \dots, N$}
        \State $P \gets (T, C, E, I)$\Comment{Prompt construction}
        \State $R \gets \theta(P)$\Comment{LLM Response}
        \State $S, F \gets \mathcal{G}(C, R)$\Comment{\texttt{is\_solve}, \texttt{feedback}}
        \If{$S$ is \textbf{True}}
            \State Break the for loop
        \Else
            \State $I \gets I + [R, F]$\Comment{Update interactions}
        \EndIf
    \EndFor
\EndFor
\end{algorithmic}
\end{algorithm}

\subsection{Models}

We employ several open-sourced LLMs known for their competitive performance on various benchmarks, including Gemma-2 9B and 27B Instruct~\cite{team2024gemma}, Llama-3.1 8B Instruct, Llama-3.3 70B Instruct~\cite{dubey2024llama}, and the Qwen-2.5 instruct models of different scales (7B, 14B, 32B, and 72B)~\cite{yang2024qwen2}. Additionally, we include proprietary closed models like GPT-4o Mini and GPT-3o Mini, given that mini models offer a good balance between performance and cost efficiency. For model inference, we implement greedy decoding to maintain deterministic outcomes. Specifically, for GPT-o3 Mini, we configure the settings to prioritize the shortest reasoning generation option. We use accuracy or solve rate as our evaluation metric to measure the correctness of the answer.

\subsection{Human Annotation}
To understand how humans play and to compare their abilities with those of LLMs, we develop a web-based interface\footnote{\url{https://huggingface.co/spaces/fhudi/textgames}} that enables human participants to engage with our games.
Through this platform, we document interactions between participants and our grading system, capturing metrics such as solve rates, the number of attempts, and the time taken to solve. These data allow us to directly compare human capabilities to those of LLMs. Each participant is asked to solve 2 to 3 different sessions. Details regarding the demographics of the annotators are available in Appendix~\ref{sec:annotator_demographic}.

\begin{figure}[!th]
    \centering
    \includegraphics[width=\linewidth]{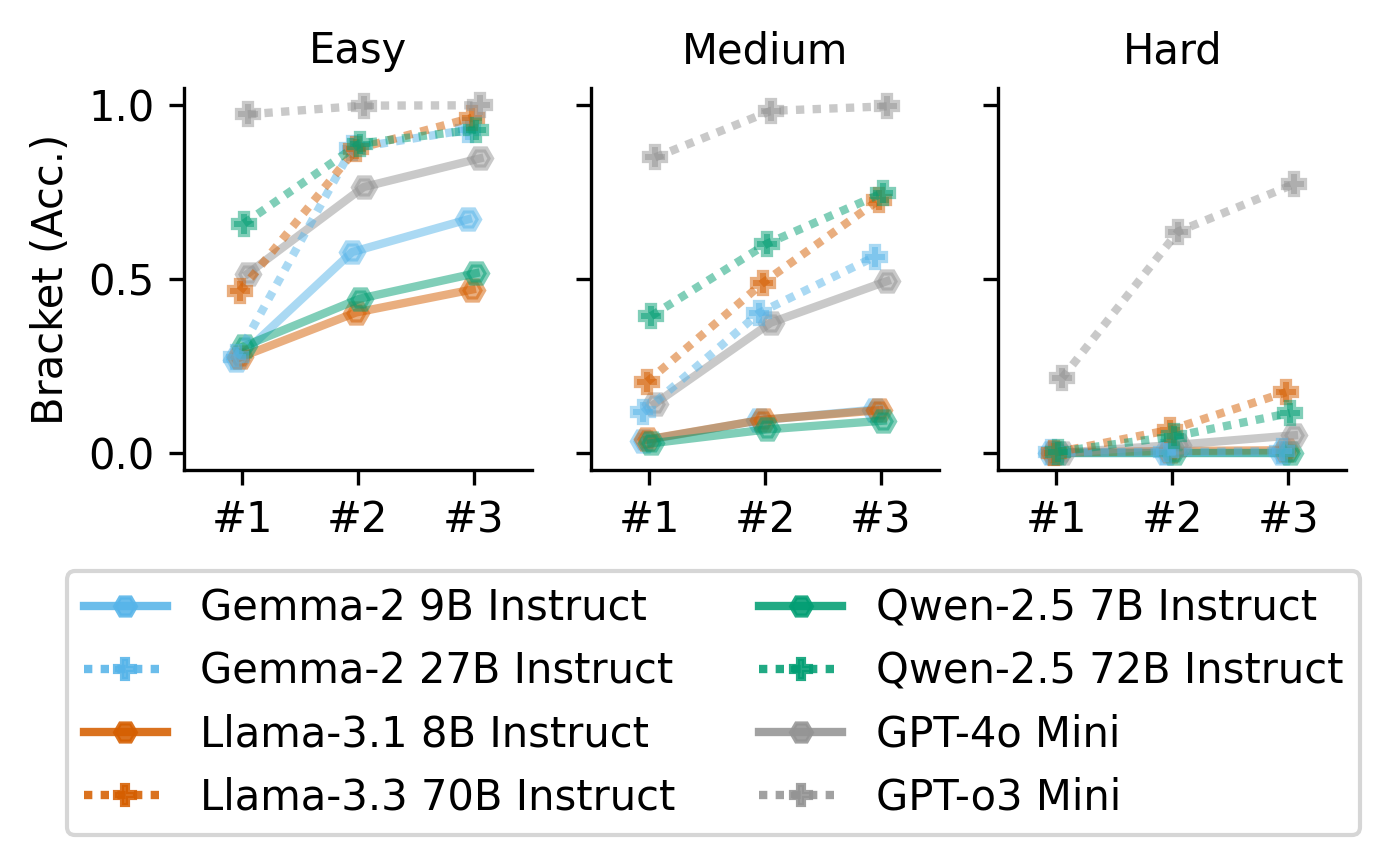}
    \caption{LLM performance on the Bracket Game in the one-shot setting, excluding GPT models. The results show that increasing the number of turns generally enhances performance. A similar trend is evident in Crossword Arranger, as shown by Figure~\ref{fig:analysis_multiturn_crossword} in the Appendix~\ref{sec:multi_turn_results_vis} showing illustrations from all games}
    \label{fig:analysis_multiturn_bracket}
\end{figure}

\section{Results and Analysis}

Our findings indicate that our benchmark poses a considerable challenge for LLMs. Even at the easiest difficulty level, the majority of models struggle to solve the games. An exception is the highly capable GPT-o3 Mini, which succeeds on only a subset of the games. This highlights the persistent difficulty of our benchmark for LLMs, highlighting areas where further advancements are needed.


\begin{table*}[!ht]
\centering
\resizebox{.8\textwidth}{!}{
\begin{tabular}{lrrr|rrr|rrr}
\toprule
 & \multicolumn{3}{c|}{\bfseries Easy} & \multicolumn{3}{c|}{\bfseries Medium} & \multicolumn{3}{c}{\bfseries Hard} \\
\multicolumn{1}{l}{\textbf{Model} \hskip4em Turn \#} & \#1 & \#2 & \#3 & \#1 & \#2 & \#3 & \#1 & \#2 & \#3 \\
\midrule
Gemma-2 9B Instruct & 39.8 & 58.0 & 64.1 & 12.0 & 19.7 & 25.2 & 2.5 & 3.6 & 4.2 \\
Gemma-2 27B Instruct & 50.7 & 77.0 & 82.4 & 18.9 & 37.5 & 46.7 & 4.1 & 7.1 & 9.3 \\
Llama-3.1 8B Instruct & 40.8 & 52.4 & 58.6 & 11.2 & 16.2 & 18.9 & 1.2 & 2.1 & 2.6 \\
Llama-3.3 70B Instruct & 55.8 & \underline{78.7} & \underline{86.4} & 23.9 & \underline{43.0} & \underline{56.6} & 5.1 & \underline{10.2} & \underline{15.3} \\
Qwen-2.5 7B Instruct & 30.6 & 44.6 & 52.5 & 8.4 & 14.2 & 18.5 & 1.2 & 1.8 & 2.3 \\
Qwen-2.5 72B Instruct & \underline{60.7} & 75.4 & 81.5 & \underline{26.9} & 40.3 & 49.3 & 3.9 & 7.8 & 11.1 \\
GPT-4o Mini & 59.6 & 74.3 & 79.0 & 22.1 & 37.6 & 45.3 & \underline{6.3} & 9.4 & 11.6 \\
GPT-o3 Mini & \bfseries 96.5 & \bfseries 98.9 & \bfseries 99.4 & \bfseries 87.2 & \bfseries 96.2 & \bfseries 97.4 & \bfseries 45.1 & \bfseries 69.5 & \bfseries 78.0 \\
\bottomrule
\end{tabular}
}
\caption{Average solve rate (\%) for multi-turn 1D Puzzles.}
\label{results_avg_3turns_1d}
\end{table*}

\begin{table*}[!ht]
\centering
\resizebox{.8\textwidth}{!}{
\begin{tabular}{lrrr|rrr|rrr}
\toprule
 & \multicolumn{3}{c|}{\bfseries Easy} & \multicolumn{3}{c|}{\bfseries Medium} & \multicolumn{3}{c}{\bfseries Hard} \\
\multicolumn{1}{l}{\textbf{Model} \hskip4em Turn \#} & \#1 & \#2 & \#3 & \#1 & \#2 & \#3 & \#1 & \#2 & \#3 \\
\midrule
Gemma-2 9B Instruct & 19.1 & 30.4 & 36.7 & 2.9 & 5.1 & 7.9 & 0.3 & 0.8 & 1.6 \\
Gemma-2 27B Instruct & 21.4 & 36.0 & 44.4 & 5.9 & 9.3 & 13.5 & 0.9 & 1.6 & 2.6 \\
Llama-3.1 8B Instruct & 6.6 & 12.7 & 20.7 & 1.4 & 2.0 & 3.7 & 0.1 & 0.3 & 0.7 \\
Llama-3.3 70B Instruct & 23.2 & 38.0 & 48.4 & 4.0 & 7.3 & 10.7 & \underline{0.9} & \underline{2.2} & 3.4 \\
Qwen-2.5 7B Instruct & 12.5 & 20.7 & 26.3 & 2.6 & 4.8 & 6.5 & 0.3 & 0.7 & 1.3 \\
Qwen-2.5 72B Instruct & \underline{28.2} & \underline{43.1} & \underline{51.1} & \underline{10.4} & \underline{15.7} & \underline{19.1} & 0.9 & 2.1 & \underline{3.4} \\
GPT-4o Mini & 20.7 & 34.7 & 40.9 & 5.1 & 9.5 & 12.4 & 0.7 & 1.8 & 2.6 \\
GPT-o3 Mini & \bfseries 87.8 & \bfseries 94.9 & \bfseries 98.8 & \bfseries 66.5 & \bfseries 80.7 & \bfseries 86.9 & \bfseries 20.6 & \bfseries 40.5 & \bfseries 48.6 \\
\bottomrule
\end{tabular}
}
\caption{Average solve rate (\%) for multi-turn 2D Puzzles.}
\label{results_avg_3turns_2d}
\end{table*}


\begin{table*}[!th]
\centering
\resizebox{.94\textwidth}{!}{
\begin{tabular}{llrrr|rrr|rrr}
\toprule
 && \multicolumn{3}{c|}{\textbf{1\textsuperscript{st} Turn Solve Rate (\%)}} & \multicolumn{3}{c|}{\textbf{Avg. Attempts}} & \multicolumn{3}{c}{\textbf{Avg. Time to Solve (s)}} \\
 && Easy & Medium & Hard & Easy & Medium & Hard & Easy & Medium & Hard \\
\midrule
\multicolumn{11}{l}{\texttt{1D Puzzles}} \\ 
 & \textbf{Anagram Scribble} & 100.0 & 87.5 & 57.1 & 1.00 & 1.12 & 2.14 & 11.7 & 82.6 & \color[HTML]{E83E8C} 263.5 \\
 & \textbf{Password Game} & 88.9 & 100.0 & 44.4 & 1.22 & 1.00 & 1.78 & 27.2 & 44.4 & 73.4 \\
 & \textbf{Bracket Game} & 100.0 & 75.0 & 75.0 & 1.00 & 1.25 & 1.25 & 29.3 & 48.9 & 71.2 \\
 & \textbf{String Search} & 100.0 & 100.0 & 75.0 & 1.00 & 1.00 & 1.38 & 14.6 & 17.4 & 41.4 \\
\midrule
\multicolumn{11}{l}{\texttt{2D Puzzles}} \\ 
 & \textbf{Crossword Arranger} & 77.8 & 100.0 & 88.9 & 1.33 & 1.00 & 1.11 & 32.2 & \color[HTML]{E83E8C} 138.7 & \color[HTML]{E83E8C} 128.2 \\
 & \textbf{Text Sudoku} & 100.0 & 100.0 & 77.8 & 1.00 & 1.00 & 1.78 & 11.7 & 29.5 & \bfseries \color[HTML]{E83E8C} 536.3 \\
 & \textbf{Islands} & 100.0 & 100.0 & 100.0 & 1.00 & 1.00 & 1.00 & 12.2 & 25.5 & 41.4 \\
 & \textbf{Ordering Text} & 55.6 & 57.1 & 42.9 & 1.67 & 3.14 & 2.00 & 72.3 & \color[HTML]{E83E8C} 127.5 & \bfseries \color[HTML]{E83E8C} 424.3 \\
\bottomrule
\end{tabular}
}
\caption{Performance of human annotators on playing $\benchmarkname$.}
\label{results_human}
\end{table*}

\paragraph{Model Scaling Improves Performance.}

Larger models generally exhibit superior performance, particularly when comparing models within the same family (e.g., Gemma-2 9B vs. 27B Instruct), where the larger model consistently outperforms its smaller counterpart. Notably, the Gemma-2 27B Instruct model remains highly competitive despite being significantly smaller than other 70B+ baselines. Typically, larger models excel on easier tasks; however, this advantage does not necessarily extend to more challenging tasks, such as those requiring reasoning in two-dimensional coordinates. This trend is illustrated in Figure \ref{fig:model_scale} in the Appendix.

\paragraph{Multi-Turn Feedback Improves LLM Performance.}

While LLMs typically underperform on single-turn attempts, we observe noteworthy improvements when they receive feedback explaining why their previous responses were incorrect. These enhancements are most evident at the easy difficulty level. Figures~\ref{fig:analysis_multiturn_bracket} and~\ref{fig:analysis_multiturn_crossword} illustrate this positive trend, showcasing how LLMs effectively use feedback from previous interactions to engage in self-reflection and refine their subsequent outputs. A similar trend is evident in the results for various models, as shown in Table~\ref{results_avg_3turns_1d} for 1D games and Table~\ref{results_avg_3turns_2d} for 2D games.

\begin{figure}[!th]
    \centering\includegraphics[width=0.49\textwidth]{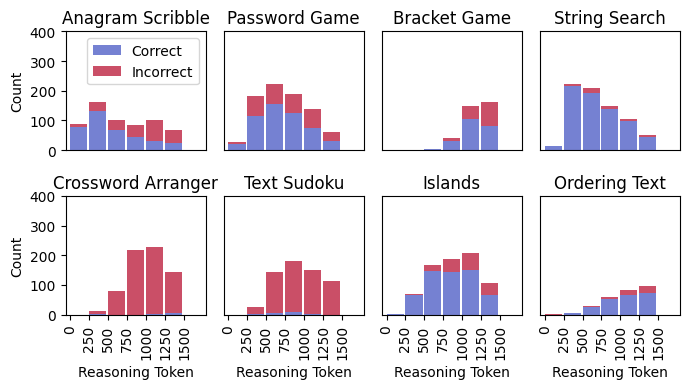}
    \caption{
    In hard games, the test-time scaling of GPT-o3 Mini displays inverse scaling behavior, with longer reasoning traces often leading to incorrect results.
    }\label{results:test-time-scaling}
\end{figure}

\paragraph{$\benchmarknameonly$ Are Solvable by Humans.}

When comparing LLM performance to human performance, we observe that humans can easily achieve full scores, especially on the easy difficulty. This is because some problems, particularly at lower difficulty levels, are arguably trivial for adult humans. On average, humans could solve all the problems within 2 attempts except for Ordering Text on the medium difficulty. This finding is particularly interesting given that recent research suggests LLMs exhibit intelligence seemingly on par with humans~\cite{achiam2023gpt}. Yet, these models struggle with tasks as simple as searching for a substring and placing a bracket around it or constructing a 2D string with a predefined number of ``islands.'' At higher difficulty levels, we observe a decline in human performance, reflected in the lower one-turn solve rate and increased time required to solve. However, while LLMs exhibit a similar trend, most models fail to solve any hard problems, whereas humans still manage to solve them in one turn.

\paragraph{Misaligned Difficulty Perception between LLMs and Humans.}

The ``Islands'' and ``String Search'' games are among the easiest problems for humans; even at the hardest difficulty, humans typically solve them in fewer than two turns, making them some of the fastest problems to complete. In contrast, LLMs struggle significantly with these tasks, generally exhibiting subpar performance. This highlights a discrepancy in difficulty perception between humans and LLMs and sheds light on the fundamental differences in how humans and LLMs approach constrained puzzle-solving.

\paragraph{Inverse-Scaling on Reasoning Length and Performance} 
Previous studies have generally shown that longer reasoning sequences enhance performance. Interestingly, this pattern is not evident in GPT-o3 Mini (Figure~\ref{results:test-time-scaling}). We observe that GPT-o3 Mini tends to produce incorrect answers more frequently with extended reasoning tokens, particularly in the Bracket Game, Islands, and Ordering Text. Although GPT-o3 Mini does not disclose its reasoning process, we hypothesize that it may become confused by its own extended reasoning, resulting in overcomplicated solutions or incorrect understanding. An empirical example is illustrated by the recent DeepSeek R1 hallucination, where the system initially provided a correct answer but, after further analysis and reasoning, can be misled into an incorrect conclusion, shown in Table~\ref{tab:deepseek_hallucinated} in the Appendix~\ref{sec:longer_reasoning_confusion}.

\section{Related Work}

\paragraph{Games using LLMs.} With the advancement of LLMs, recent works examine their capabilities in playing games or assisting humans in gameplay~\cite{hu2024survey}. Classical games like Go~\cite{silver2017mastering}, chess~\cite{feng2024chessgpt}, Poker~\cite{huang2024pokergpt} have been used as initial testbeds for evaluating models' planning and decision-making abilities. More recently, more works have explored other genres for more dynamic and complex situations like text-based games~\cite{xiao2024llms, AGI-BY-2028, kazemi2024boardgameqa}, communication games~\cite{guan2025richelieu,xu2025learning}, and modern strategic video games~\cite{zhang2023creative,hu2024pok,qi2024civrealm,rao2024collaborative,ma2025large}. 
In comparison, \benchmarknameonly{} takes inspiration from real-life text puzzle games and emphasizes evaluating LLM's capabilities in simple logic reasoning. Additionally, each game come with different level of difficulty for assessing the models' robustness.

\paragraph{Text-based Reasoning.} Text-based reasoning has been extensively studied across various domains, including commonsense reasoning~\cite{rajani2019explain,bhargava2022commonsense,zhao2023large}, mathematical reasoning~\cite{patel2021nlp,zhao2022multihiertt,lu2023mathvista}, logical reasoning~\cite{pan2023logic}, causal reasoning~\cite{wang2024causalbench,jin2024cladder}, and agent-based reasoning~\cite{motwani2024malt}. While existing benchmarks assess different aspects of reasoning, they often evaluate these abilities in isolation. In contrast, $\benchmarknameonly$ assesses LLMs' capacity for integrating multiple reasoning skills, offering a richer evaluation of model strengths and weaknesses.
~
\section{Conclusion}
We present $\benchmarknameonly$, a text-based puzzle game benchmark designed to evaluate the diverse reasoning abilities of LLMs, including pattern recognition, spatial awareness, arithmetic, and logical reasoning. In addition to only evaluating single-turn solve rate, our evaluation system also implement feedback in multi-turn gameplay settings and test whether models improve through self-reflection. Results show that while LLMs proficiently solve most easy and medium-level problems, they encounter significant challenges with more difficult tasks that demand comprehensive reasoning. In contrast, humans can solve all tasks given sufficient time. We show significant performance improvement with multi-turn prediction via self-reflection. We hope $\benchmarknameonly$ could contribute to uncovering and analyzing the weaknesses of LLMs in complex reasoning tasks.


\section*{Limitations}
In this paper, we focus our investigation by not exhaustively evaluating every possible model, owing to resource constraints. Instead, our primary objective is to develop a benchmark that serves as a platform for future research exploration on reasoning.

\section*{Ethical Considerations}
In conducting our research, which focuses on evaluating LLMs for complex reasoning tasks, we are committed to upholding the highest standards of transparency and fairness in all aspects of our data collection and evaluation processes. We ensure that the methodologies and criteria used for assessment are clearly documented and unbiased, promoting fair comparisons across different models. Our commitment to these principles aims to foster trust and accountability in our research outcomes.

\section*{Acknowledgments}
We extend our gratitude to members of NAIST NLP Laboratory for their support of our project, with spacial thanks to Huayang Li for the insightful discussions and valuable recommendations.

\bibliography{custom}

\appendix


\section{GPU computation usage and Hyperparameters}

We employ NVIDIA GPUs, RTX A6000 (48GB) and RTX 6000 (48GB), to run inference for the whole open model which took us the equivalent of $\sim$650 GPU hours. We apply the default parameters as defined from each models respective HuggingFace's page for all of our experiments. To allow reproducibility, we use greedy decoding, i.e. by setting parameter \texttt{do\_sample} to \texttt{False}.

\section{Dataset License}
We will release our dataset under the open-source CC-BY-SA 4.0 license, facilitating redistribution for future research.

\section{Attribution}
\label{sec:attribution}

The icon images on Figure~\ref{fig:game} is taken from \url{https://flaticon.com}. They are freely for personal and commercial use with attribution.


\section{Model Scale Improvement}
\label{sec:model_scale}

\begin{figure}[!th]
    \centering
    \includegraphics[width=\linewidth]{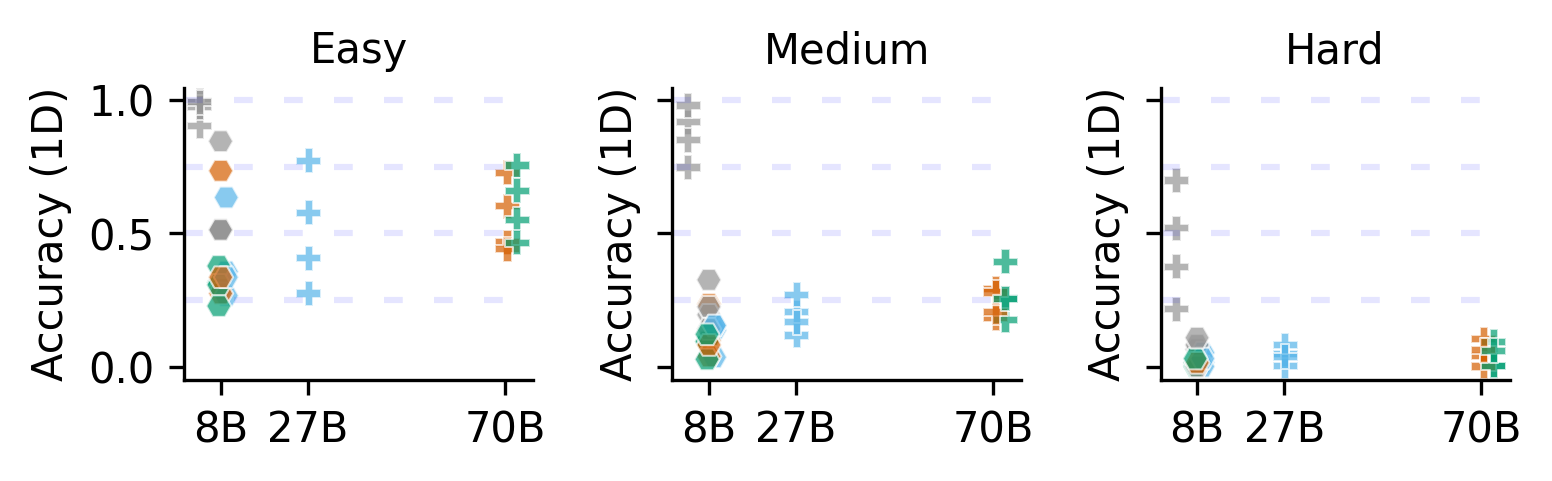}
    \includegraphics[width=\linewidth]{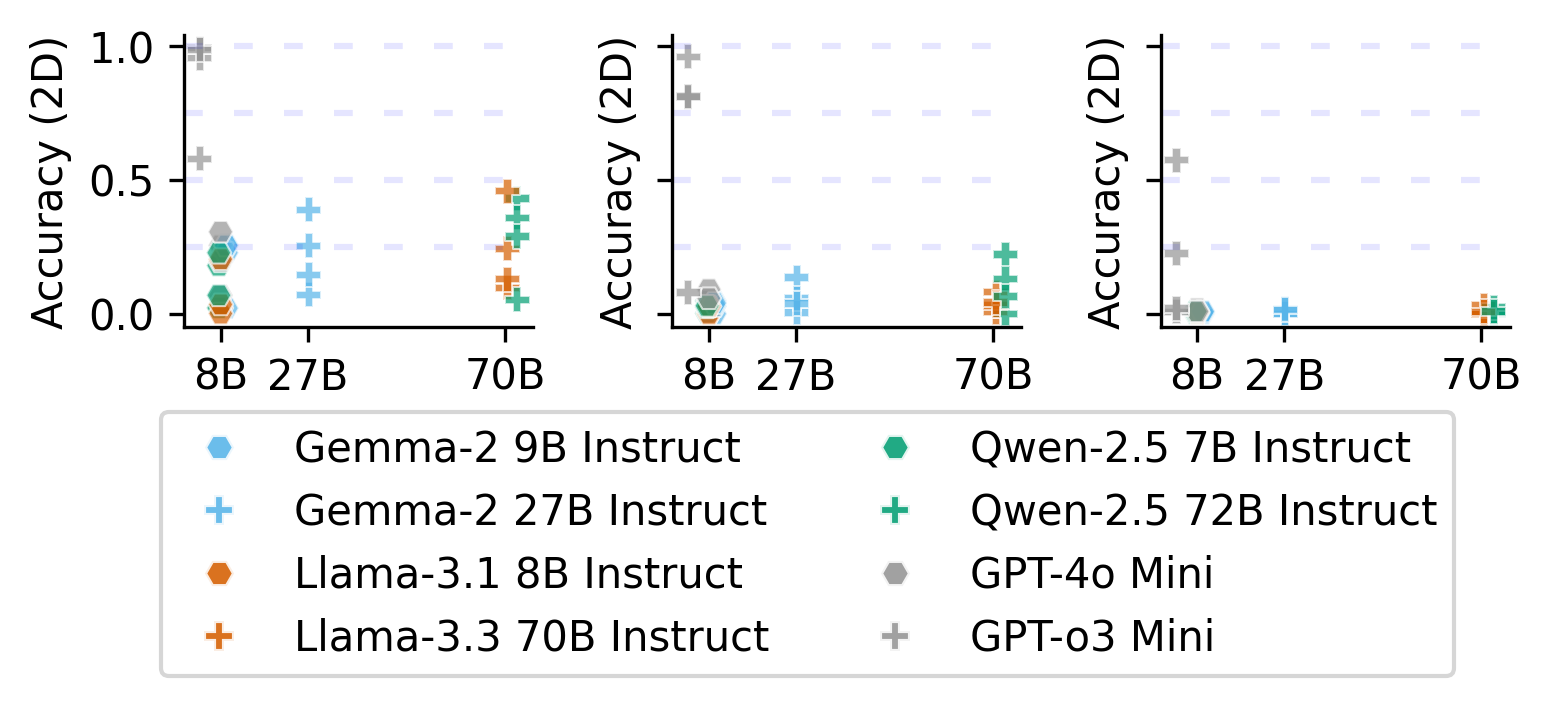}
    \caption{Model scaling improves easier tasks.}
    \label{fig:model_scale}
\end{figure}

Figure \ref{fig:model_scale} illustrates how the scale of the model impacts performance, with variations depending on task difficulty.

\section{Annotator Demographic}
\label{sec:annotator_demographic}

There are 4 annotators, within the age range of 25-35 years old, voluntarily participating in our experiments.
All annotators are from Computer Science background with a degree of magisterial or doctoral.
All 4 annotators are fluent English speakers from Asia-based origins with experience living in English-speaking countries and have been using English for more than 15 years. All annotators have given consent for using, releasing and redistributing their annotations.

\vfill
\eject

\section{Multi-turn Results Visualization}
\label{sec:multi_turn_results_vis}
\begin{figure}[!th]
    \centering
    \includegraphics[width=\linewidth,trim={0 89px 0  0px},clip]{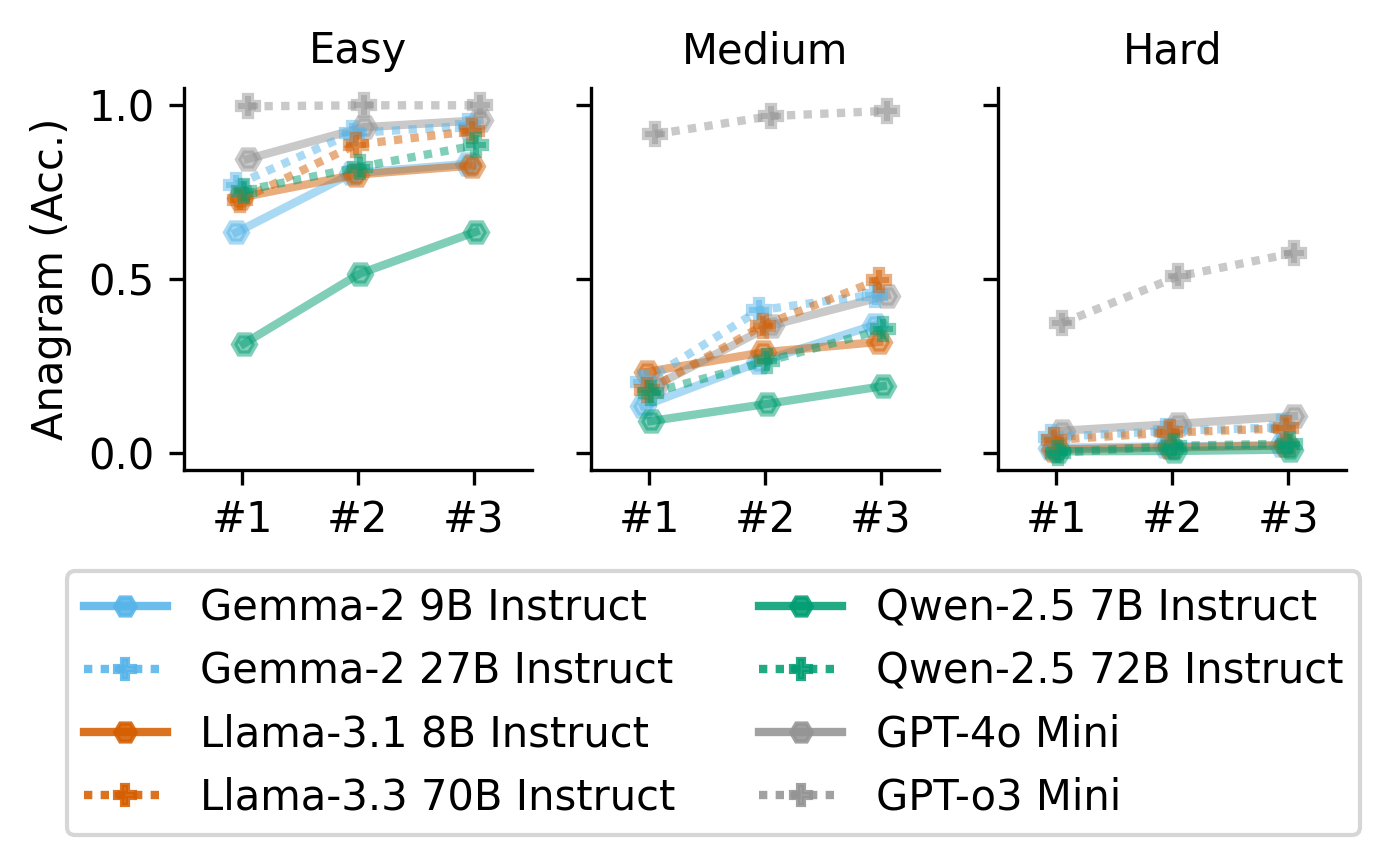}
    \includegraphics[width=\linewidth,trim={0 89px 0 18px},clip]{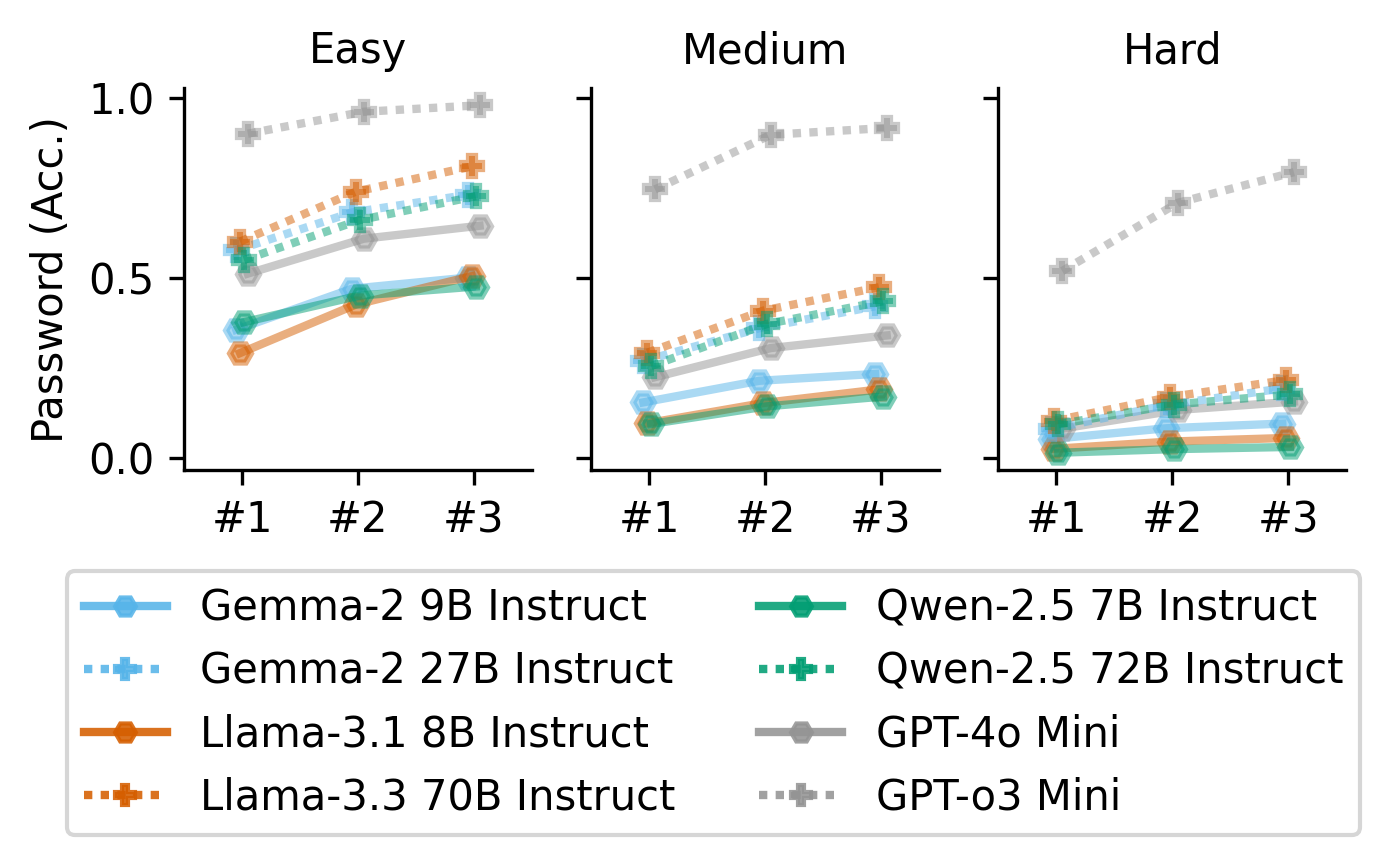}
    \includegraphics[width=\linewidth,trim={0 89px 0 18px},clip]{imgs/analysis_multiturn_bracket.png}
    \includegraphics[width=\linewidth,trim={0 89px 0 18px},clip]{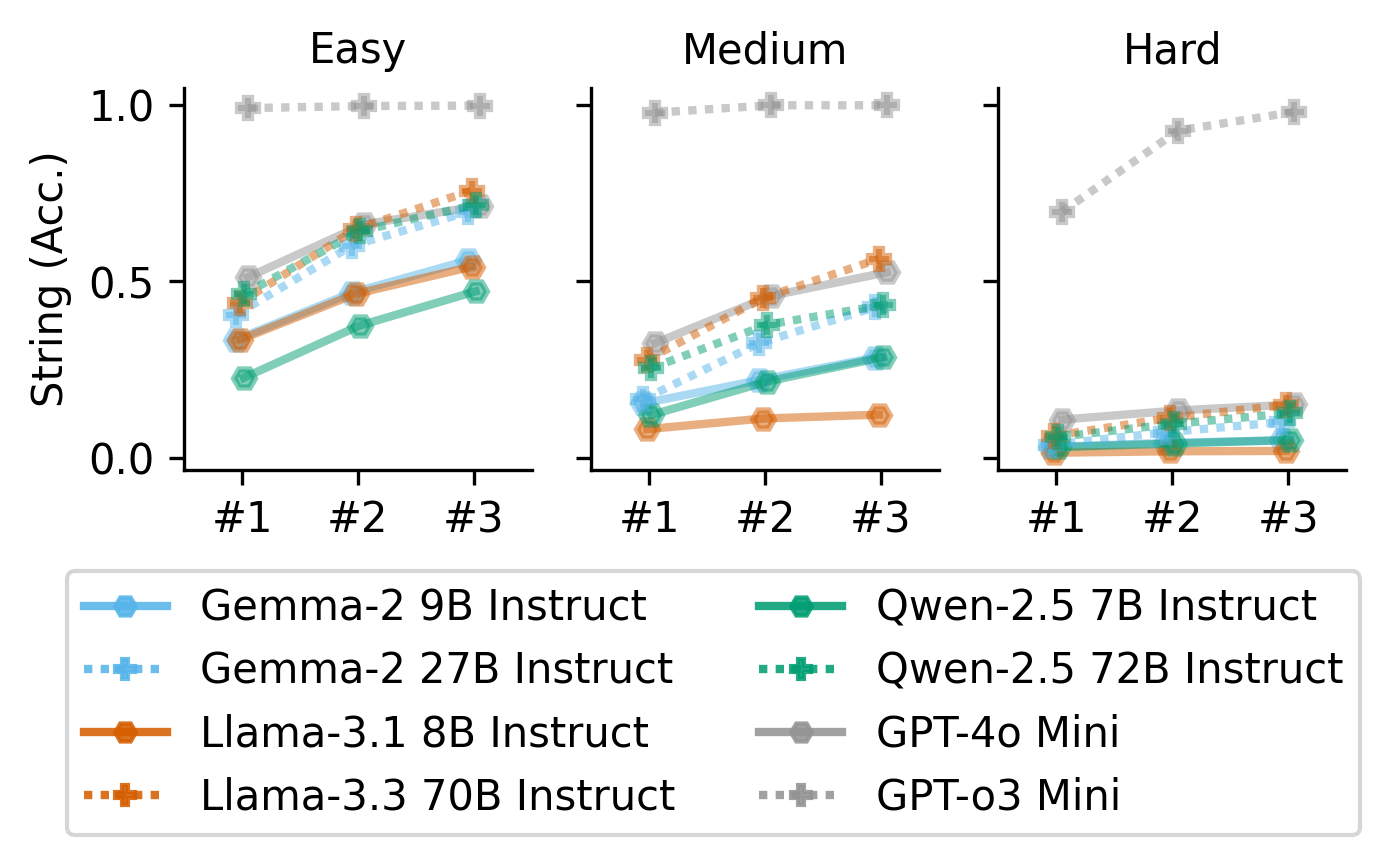}
    \includegraphics[width=\linewidth,trim={0 89px 0 18px},clip]{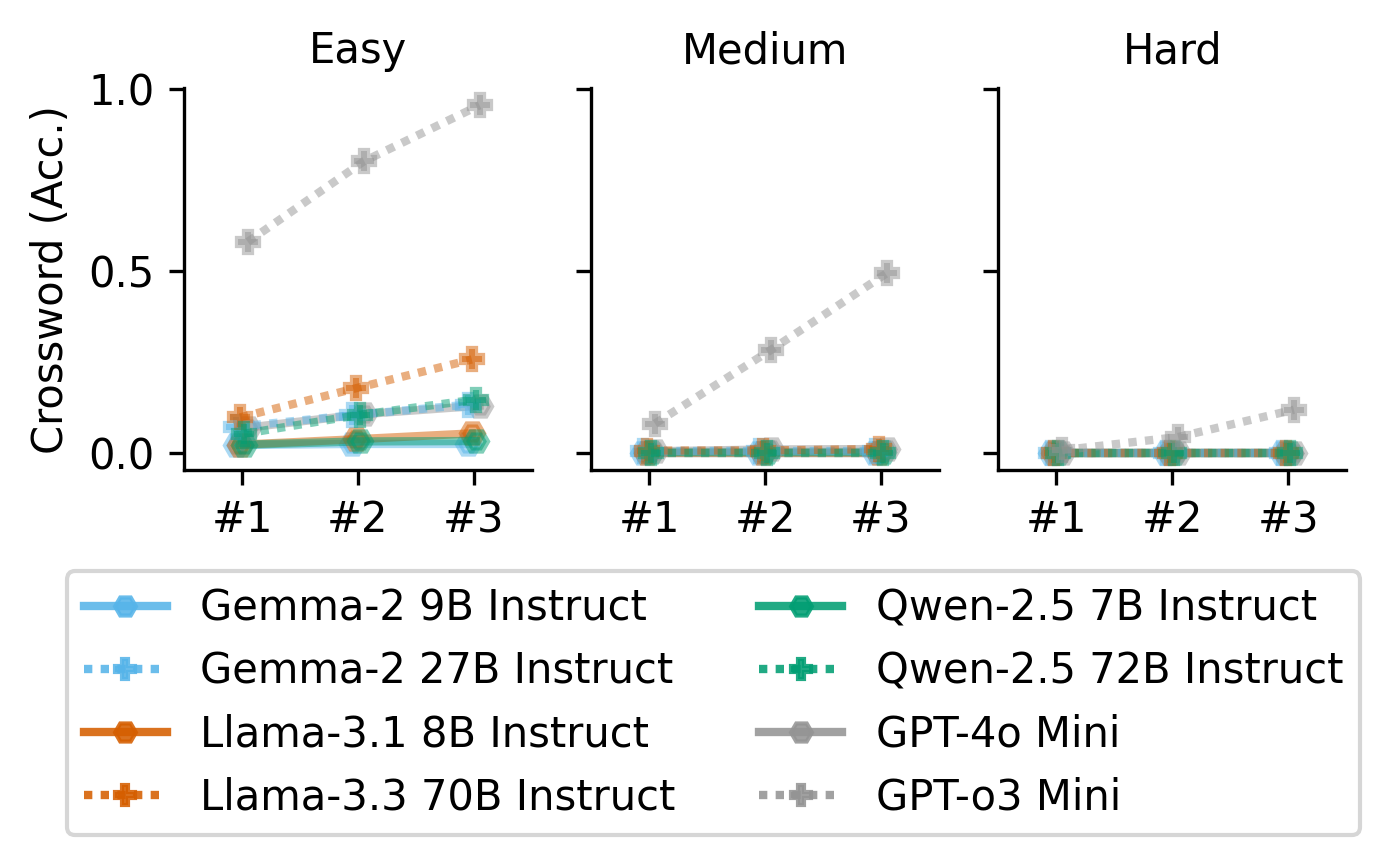}
    \includegraphics[width=\linewidth,trim={0 89px 0 18px},clip]{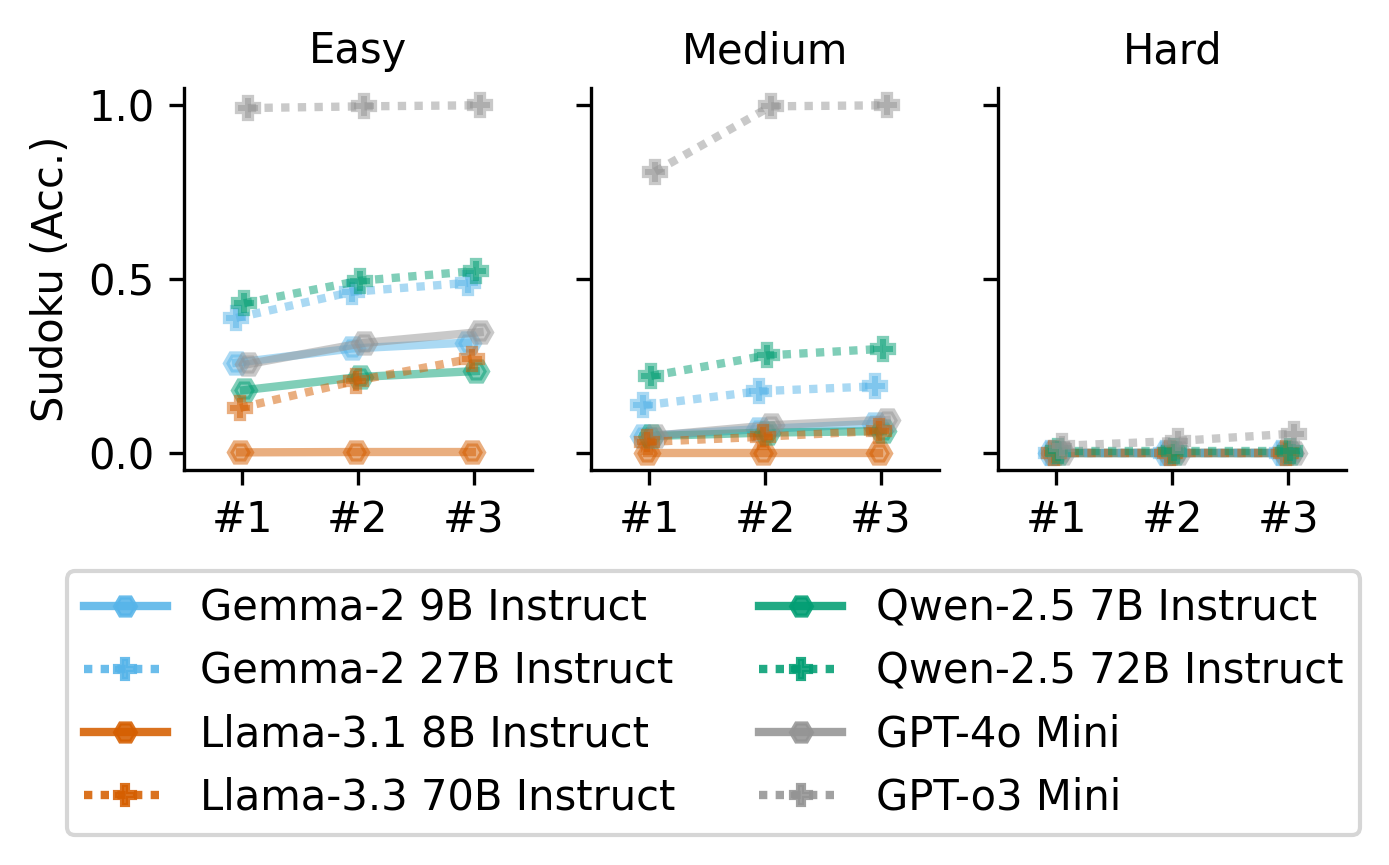}
    \includegraphics[width=\linewidth,trim={0 89px 0 18px},clip]{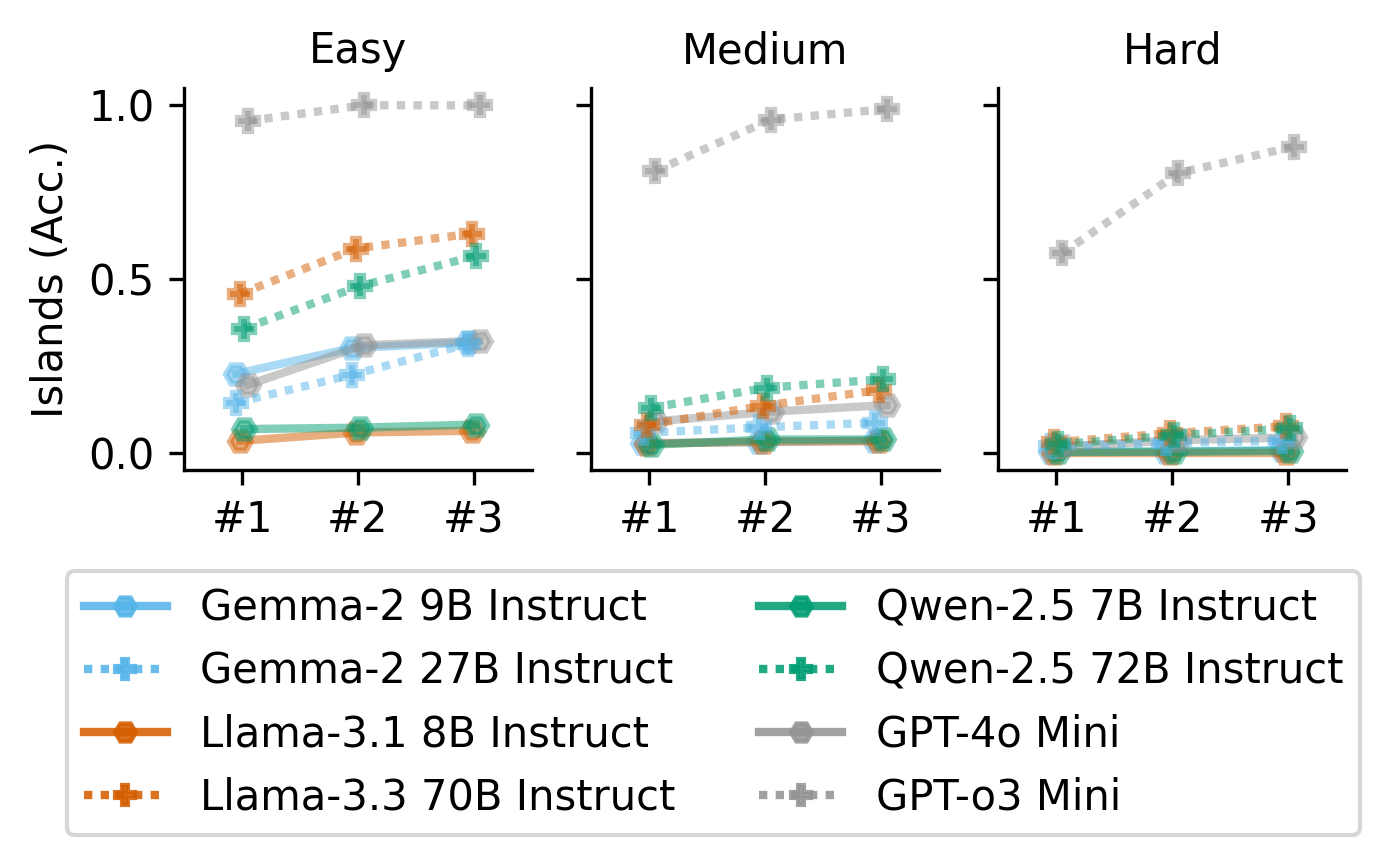}
    \includegraphics[width=\linewidth,trim={0  0px 0 18px},clip]{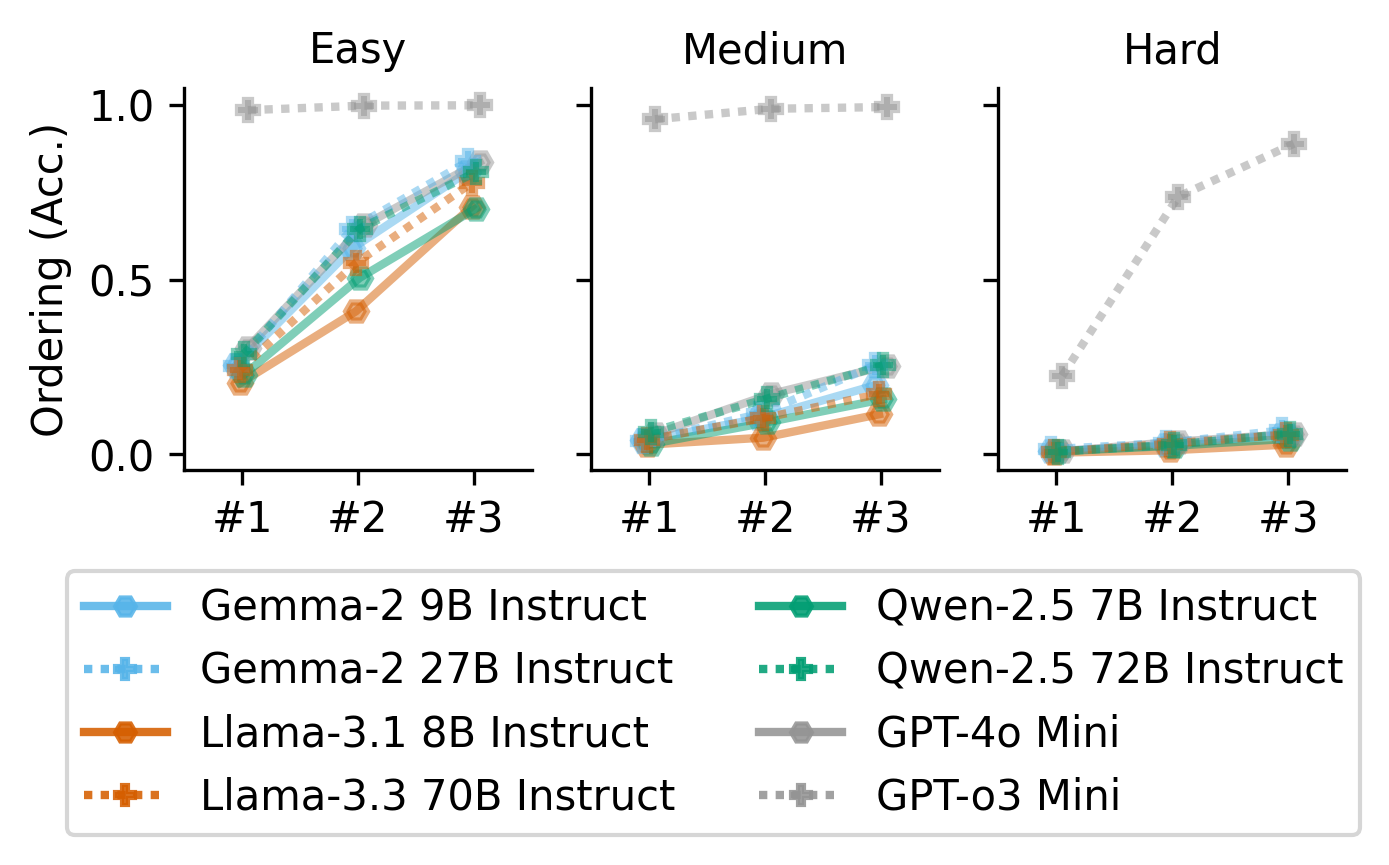}
    \caption{Multi-turn with feedback based on the performance for each game puzzle.}
    \label{fig:analysis_multiturn_crossword}
\end{figure}



\onecolumn
\section{Complete Experiment Results}

\begin{figure*}[!th]
    \centering
    \includegraphics[width=\textwidth]{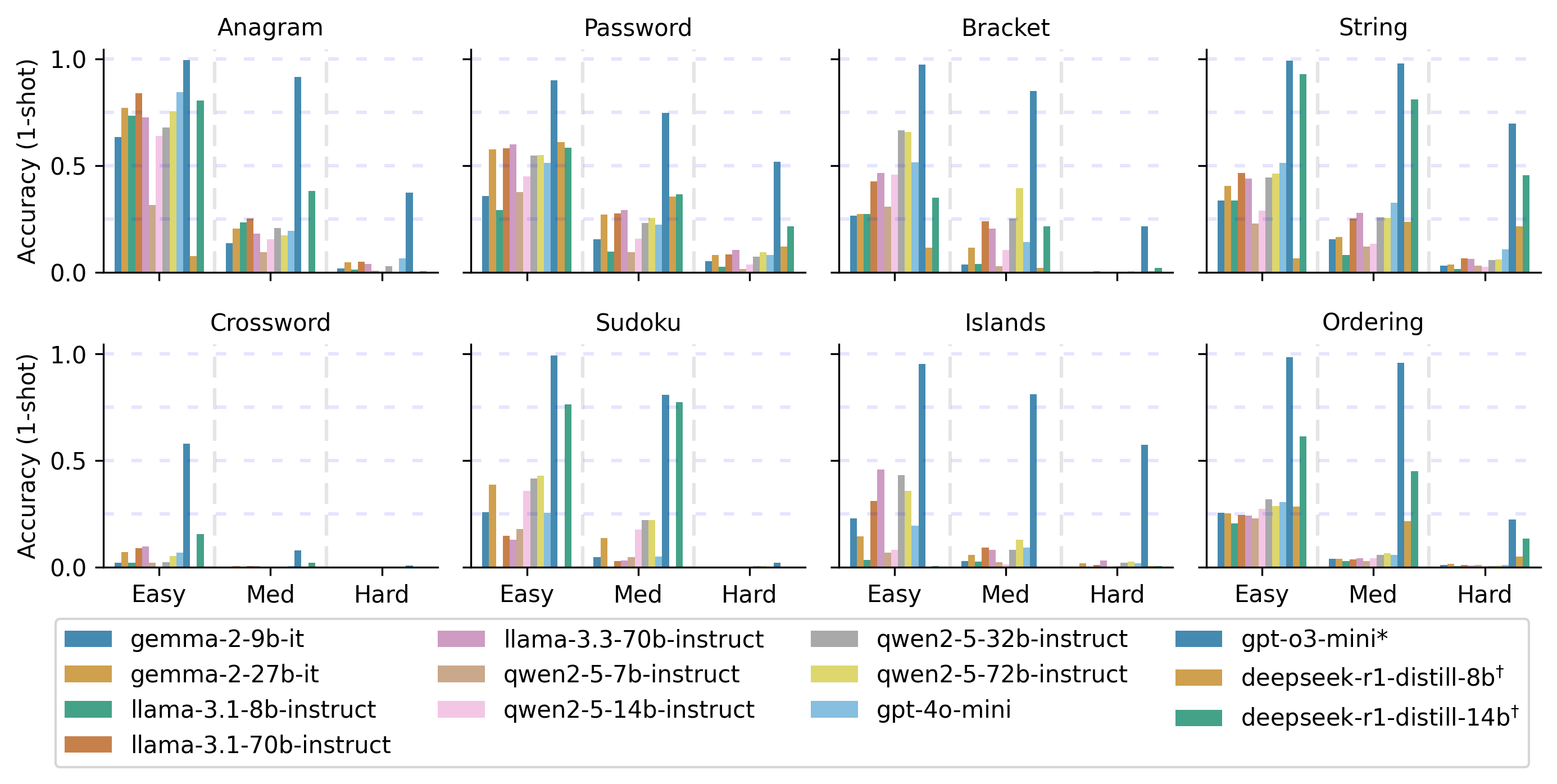}
    \caption{Complete LLMs results against \benchmarkname. \text{*}We present zero-shot results as reference}\label{results:1_shot_all}
\end{figure*}

We report the complete results that include all the models we evaluated on, as illustrated in Figure~\ref{results:1_shot_all}. The numerical results of these models can be found in Table~\ref{results_1d_all} and Table~\ref{results_2d_all}, with the respective Zero-Shot setting performance in Table~\ref{results_1d_all_zero_shot} and Table~\ref{results_2d_all_zero_shot}.
We also report the performance of multi-turn settings for each game: Anagram Scribble in Table~\ref{results_3turns_anagram}, Password Games in Table~\ref{results_3turns_password}, Bracket Game in Table~\ref{results_3turns_bracket}, String Search in Table~\ref{results_3turns_string}, Crossword Arranger in Table~\ref{results_3turns_crossword}, Text Sudoku in Table~\ref{results_3turns_sudoku}, Islands in Table~\ref{results_3turns_islands}, and Ordering Text in Table~\ref{results_3turns_ordering}. $^\dagger$Indicates first 20\% of dataset only.

\begin{table*}[!ht]
\centering
\resizebox{.9\textwidth}{!}{
\begin{tabular}{lrrr|rrr|rrr|rrr}
\toprule
 & \multicolumn{3}{c|}{\bfseries Anagram} & \multicolumn{3}{c|}{\bfseries Password} & \multicolumn{3}{c|}{\bfseries Bracket} & \multicolumn{3}{c}{\bfseries String} \\
 & Easy & Med & Hard & Easy & Med & Hard & Easy & Med & Hard & Easy & Med & Hard \\
\midrule
Gemma-2 9B Instruct & 63.4 & 13.6 & 1.6 & 35.6 & 15.4 & 5.2 & 26.6 & 3.5 & 0.0 & 33.5 & 15.4 & 3.0 \\
Gemma-2 27B Instruct & 77.1 & 20.4 & 4.7 & 57.7 & 26.9 & 8.0 & 27.4 & 11.6 & 0.1 & 40.6 & 16.6 & 3.6 \\
Llama-3.1 8B Instruct & 73.4 & 23.4 & 1.1 & 29.1 & 9.6 & 2.5 & 27.4 & 3.8 & 0.0 & 33.5 & 8.1 & 1.4 \\
Llama-3.1 70B Instruct & 84.0 & 25.1 & 5.0 & 58.1 & 27.5 & 8.3 & 42.7 & 23.9 & 0.3 & 46.6 & 25.3 & 6.5 \\
Llama-3.3 70B Instruct & 72.7 & 18.1 & 3.9 & 60.1 & 29.2 & 10.3 & 46.5 & 20.4 & 0.0 & 44.0 & 27.8 & 6.2 \\
Qwen-2.5 7B Instruct & 31.5 & 9.3 & 0.6 & 37.7 & 9.4 & 1.4 & 30.6 & 2.7 & 0.0 & 22.7 & 12.1 & 3.0 \\
Qwen-2.5 14B Instruct & 64.0 & 15.4 & 0.5 & 44.9 & 15.8 & 3.7 & 45.7 & 10.4 & 0.0 & 28.8 & 13.3 & 2.6 \\
Qwen-2.5 32B Instruct & 67.9 & 20.8 & 2.9 & 54.8 & 23.1 & 7.3 & \underline{66.5} & 25.1 & 0.1 & 44.4 & 25.7 & 5.8 \\
Qwen-2.5 72B Instruct & 75.4 & 17.4 & 0.2 & 55.0 & 25.5 & 9.3 & 65.9 & \underline{39.3} & 0.3 & 46.4 & 25.4 & 5.9 \\
GPT-4o Mini & \underline{84.5} & 19.4 & \underline{6.5} & 51.2 & 22.4 & 8.1 & 51.5 & 14.1 & 0.0 & 51.2 & 32.5 & 10.8 \\
\midrule
GPT-o3 Mini* & \bfseries 99.6 & \bfseries 91.6 & \bfseries 37.4 & \bfseries 90.1 & \bfseries 74.6 & \bfseries 51.9 & \bfseries 97.3 & \bfseries 84.9 & \bfseries 21.5 & \bfseries 99.2 & \bfseries 97.9 & \bfseries 69.8 \\
DeepSeek-R1-Distill 8B$^\dagger$ & 7.5 & 0.0 & 0.0 & \underline{61.0} & 35.5 & 12.0 & 11.5 & 2.0 & 0.5 & 6.5 & 23.5 & 21.5 \\
DeepSeek-R1-Distill 14B$^\dagger$ & 80.5 & \underline{38.0} & 0.5 & 58.5 & \underline{36.5} & \underline{21.5} & 35.0 & 21.5 & \underline{2.0} & \underline{93.0} & \underline{81.0} & \underline{45.5} \\
\bottomrule
\end{tabular}
}
\caption{Complete Average Results (\%) for 1D Puzzles. \text{*}Zero-shot results as reference}
\label{results_1d_all}
\end{table*}

\begin{table*}[!ht]
\centering
\resizebox{.9\textwidth}{!}{
\begin{tabular}{lrrr|rrr|rrr|rrr}
\toprule
 & \multicolumn{3}{c|}{\bfseries Crossword} & \multicolumn{3}{c|}{\bfseries Sudoku} & \multicolumn{3}{c|}{\bfseries Islands} & \multicolumn{3}{c}{\bfseries Ordering} \\
 & Easy & Med & Hard & Easy & Med & Hard & Easy & Med & Hard & Easy & Med & Hard \\
\midrule
Gemma-2 9B Instruct & 2.1 & 0.0 & 0.0 & 25.9 & 4.8 & 0.0 & 22.8 & 2.9 & 0.3 & 25.5 & 4.0 & 0.9 \\
Gemma-2 27B Instruct & 7.1 & 0.5 & 0.0 & 38.8 & 13.6 & 0.0 & 14.5 & 5.8 & 1.9 & 25.4 & 3.8 & 1.5 \\
Llama-3.1 8B Instruct & 2.2 & 0.0 & 0.0 & 0.1 & 0.0 & 0.0 & 3.5 & 2.7 & 0.1 & 20.4 & 2.8 & 0.5 \\
Llama-3.1 70B Instruct & 8.9 & 0.4 & \underline{0.1} & 14.7 & 2.8 & 0.0 & 31.2 & 9.1 & 1.0 & 24.5 & 3.6 & 1.0 \\
Llama-3.3 70B Instruct & 9.7 & 0.6 & 0.0 & 12.9 & 3.1 & 0.0 & \underline{45.8} & 8.1 & \underline{3.1} & 24.3 & 4.2 & 0.7 \\
Qwen-2.5 7B Instruct & 2.2 & 0.2 & 0.0 & 18.0 & 4.8 & 0.0 & 6.9 & 2.4 & 0.3 & 22.8 & 3.0 & 0.9 \\
Qwen-2.5 14B Instruct & 0.0 & 0.0 & 0.0 & 35.9 & 17.7 & 0.3 & 8.2 & 1.2 & 0.5 & 27.4 & 4.1 & 0.5 \\
Qwen-2.5 32B Instruct & 2.3 & 0.0 & 0.0 & 41.7 & 22.0 & \underline{0.4} & 43.3 & 8.1 & 2.2 & 31.9 & 5.7 & 0.4 \\
Qwen-2.5 72B Instruct & 5.2 & 0.0 & 0.0 & 43.0 & 22.1 & \underline{0.4} & 35.8 & \underline{13.0} & 2.5 & 28.8 & 6.5 & 0.8 \\
GPT-4o Mini & 6.9 & 0.4 & 0.0 & 25.5 & 4.9 & 0.0 & 19.6 & 9.1 & 1.9 & 30.6 & 5.8 & 0.9 \\
\midrule
GPT-o3 Mini* & \bfseries 57.9 & \bfseries 8.0 & \bfseries 0.7 & \bfseries 99.2 & \bfseries 80.9 & \bfseries 2.0 & \bfseries 95.5 & \bfseries 81.1 & \bfseries 57.4 & \bfseries 98.6 & \bfseries 96.0 & \bfseries 22.5 \\
DeepSeek-R1-Distill 8B$^\dagger$ & 0.0 & 0.0 & 0.0 & 0.0 & 0.0 & 0.0 & 0.0 & 0.0 & 0.5 & 28.5 & 21.5 & 5.0 \\
DeepSeek-R1-Distill 14B$^\dagger$ & \underline{15.5} & \underline{2.0} & 0.0 & \underline{76.5} & \underline{77.5} & 0.0 & 0.5 & 0.0 & 0.5 & \underline{61.5} & \underline{45.0} & \underline{13.5} \\
\bottomrule
\end{tabular}
}
\caption{Complete Average Results (\%) for 2D Puzzles. \text{*}Zero-shot results as reference}
\label{results_2d_all}
\end{table*}

\begin{table*}[!ht]
\centering
\resizebox{.9\textwidth}{!}{
\begin{tabular}{lrrr|rrr|rrr|rrr}
\toprule
 & \multicolumn{3}{c|}{\bfseries Anagram} & \multicolumn{3}{c|}{\bfseries Password} & \multicolumn{3}{c|}{\bfseries Bracket} & \multicolumn{3}{c}{\bfseries String} \\
 & Easy & Med & Hard & Easy & Med & Hard & Easy & Med & Hard & Easy & Med & Hard \\
\midrule
Gemma-2 9B Instruct & 77.7 & 14.2 & 1.0 & 36.4 & 13.8 & 4.6 & 2.3 & 0.3 & 0.0 & 37.7 & 20.2 & 5.3 \\
Gemma-2 27B Instruct & \underline{88.0} & 31.8 & 3.1 & 45.9 & 18.7 & 6.6 & 29.8 & 5.4 & 0.0 & 44.1 & 23.7 & 4.4 \\
Llama-3.1 8B Instruct & 56.3 & 7.5 & 0.1 & 31.8 & 7.2 & 1.3 & 0.1 & 0.0 & 0.0 & 17.2 & 6.9 & 0.3 \\
Llama-3.1 70B Instruct & 69.5 & 27.7 & 2.5 & 45.1 & 19.0 & 5.6 & 23.0 & 6.0 & 0.4 & 45.9 & 25.4 & 5.8 \\
Llama-3.3 70B Instruct & 77.4 & 30.9 & 3.2 & 47.5 & 20.0 & 5.9 & 34.0 & 14.0 & 0.6 & 45.0 & 26.3 & 5.4 \\
Qwen-2.5 7B Instruct & 8.2 & 0.2 & 0.0 & 32.9 & 9.2 & 1.3 & 6.4 & 1.1 & 0.0 & 25.4 & 11.1 & 1.4 \\
Qwen-2.5 14B Instruct & 23.1 & 7.2 & 0.6 & 34.0 & 12.0 & 2.4 & 22.9 & 3.7 & 0.0 & 32.7 & 14.4 & 1.8 \\
Qwen-2.5 32B Instruct & 68.7 & 16.7 & 2.3 & 47.2 & 20.2 & 6.0 & 47.6 & 15.8 & 0.3 & 41.4 & 27.0 & 6.1 \\
Qwen-2.5 72B Instruct & 30.0 & 0.4 & 0.1 & 50.0 & 21.8 & 8.3 & \underline{58.3} & 19.9 & 0.0 & 46.7 & 26.7 & 7.0 \\
GPT-4o Mini & 79.9 & 26.7 & \underline{5.7} & 46.9 & 18.9 & 6.5 & 26.6 & 7.2 & 0.0 & 39.8 & 28.1 & 7.6 \\
GPT-o3 Mini & \bfseries 99.6 & \bfseries 91.6 & \bfseries 37.4 & \bfseries 90.1 & \bfseries 74.6 & \bfseries 51.9 & \bfseries 97.3 & \bfseries 84.9 & \bfseries 21.5 & \bfseries 99.2 & \bfseries 97.9 & \bfseries 69.8 \\
\midrule
DeepSeek-R1-Distill 8B$^\dagger$ & 7.5 & 0.0 & 0.0 & 61.0 & 35.5 & 12.0 & 11.5 & 2.0 & 0.5 & 6.5 & 23.5 & 21.5 \\
DeepSeek-R1-Distill 14B$^\dagger$ & 81.0 & \underline{35.0} & 2.0 & \underline{67.0} & \underline{45.5} & \underline{16.5} & 51.5 & \underline{38.5} & \underline{5.0} & \underline{94.5} & \underline{68.5} & \underline{58.0} \\
\bottomrule
\end{tabular}
}
\caption{Complete Average Results (\%) for 1D Puzzles (Zero-Shot).}
\label{results_1d_all_zero_shot}
\end{table*}

\begin{table*}[!ht]
\centering
\resizebox{\textwidth}{!}{
\begin{tabular}{lrrr|rrr|rrr|rrr}
\toprule
 & \multicolumn{3}{c|}{\bfseries Crossword} & \multicolumn{3}{c|}{\bfseries Sudoku} & \multicolumn{3}{c|}{\bfseries Islands} & \multicolumn{3}{c}{\bfseries Ordering} \\
 & Easy & Med & Hard & Easy & Med & Hard & Easy & Med & Hard & Easy & Med & Hard \\
\midrule
Gemma-2 9B Instruct & 1.1 & 0.0 & 0.0 & 24.3 & 3.1 & 0.0 & 1.8 & 0.0 & 0.5 & 20.6 & 2.6 & 0.9 \\
Gemma-2 27B Instruct & 6.6 & 0.0 & 0.0 & 39.1 & 15.3 & 0.0 & 10.3 & 2.4 & 0.0 & 21.9 & 2.9 & 1.2 \\
Llama-3.1 8B Instruct & 0.0 & 0.0 & 0.0 & 0.0 & 0.0 & 0.0 & 0.0 & 1.3 & 0.0 & 0.3 & 0.0 & 0.0 \\
Llama-3.1 70B Instruct & 12.2 & 0.0 & 0.0 & 7.2 & 2.7 & 0.0 & 0.0 & 0.0 & 0.0 & 8.4 & 0.9 & 0.4 \\
Llama-3.3 70B Instruct & 5.5 & 0.1 & 0.0 & 7.1 & 1.4 & 0.0 & 1.9 & 3.2 & 0.8 & 2.1 & 0.1 & 0.0 \\
Qwen-2.5 7B Instruct & 0.0 & 0.0 & 0.0 & 0.0 & 0.0 & 0.0 & 0.0 & 0.0 & 0.0 & 22.1 & 2.0 & 0.3 \\
Qwen-2.5 14B Instruct & 1.3 & 0.0 & 0.0 & 31.1 & 15.1 & 0.3 & 0.8 & 1.3 & 0.3 & 18.6 & 2.7 & 0.8 \\
Qwen-2.5 32B Instruct & 7.3 & 0.0 & 0.0 & 34.2 & 15.5 & 0.5 & 0.0 & 0.0 & 0.0 & 27.3 & 5.2 & 0.7 \\
Qwen-2.5 72B Instruct & 0.0 & 0.0 & 0.0 & \underline{42.7} & 20.0 & 0.1 & 0.0 & 0.0 & 0.0 & 22.9 & 4.6 & 0.6 \\
GPT-4o Mini & 14.0 & \underline{4.6} & \underline{0.1} & 1.1 & 0.2 & 0.0 & \underline{31.8} & 5.2 & 0.8 & 22.6 & 2.5 & 1.0 \\
GPT-o3 Mini & \bfseries 57.9 & \bfseries 8.0 & \bfseries 0.7 & \bfseries 99.2 & \bfseries 80.9 & \bfseries 2.0 & \bfseries 95.5 & \bfseries 81.1 & \bfseries 57.4 & \bfseries 98.6 & \bfseries 96.0 & \bfseries 22.5 \\
\midrule
DeepSeek-R1-Distill 8B$^\dagger$ & 0.0 & 0.0 & 0.0 & 0.0 & 0.0 & 0.0 & 0.0 & 0.0 & 0.5 & 28.5 & 21.5 & 5.0 \\
DeepSeek-R1-Distill 14B$^\dagger$ & \underline{15.5} & 4.0 & 0.0 & 34.0 & \underline{22.0} & \underline{1.0} & 3.0 & \underline{10.0} & \underline{8.0} & \underline{65.0} & \underline{59.0} & \underline{16.5} \\
\bottomrule
\end{tabular}
}
\caption{Complete Average Results (\%) for 2D Puzzles (Zero-Shot).}
\label{results_2d_all_zero_shot}
\end{table*}



\begin{table*}[!ht]
\centering
\resizebox{.8\textwidth}{!}{
\begin{tabular}{lrrr|rrr|rrr}
\toprule
 & \multicolumn{9}{c}{\bfseries Anagram} \\
 & \multicolumn{3}{c|}{Easy} & \multicolumn{3}{c|}{Med} & \multicolumn{3}{c}{Hard} \\
\multicolumn{1}{l}{\textbf{Model} \hskip4em Turn \#} & \#1 & \#2 & \#3 & \#1 & \#2 & \#3 & \#1 & \#2 & \#3 \\
\midrule
Gemma-2 9B Instruct & 63.4 & 80.4 & 83.1 & 13.6 & 26.2 & 36.9 & 1.6 & 2.0 & 2.3 \\
Gemma-2 27B Instruct & 77.1 & 91.9 & 94.0 & 20.4 & \underline{41.1} & 45.6 & 4.7 & 6.4 & 7.5 \\
Llama-3.1 8B Instruct & 73.4 & 80.0 & 82.6 & 23.4 & 29.0 & 32.0 & 1.1 & 1.7 & 2.3 \\
Llama-3.1 70B Instruct & 84.0 & 91.1 & 93.4 & \underline{25.1} & 40.6 & 49.3 & 5.0 & 7.0 & 7.8 \\
Llama-3.3 70B Instruct & 72.7 & 88.6 & 92.5 & 18.1 & 36.7 & \underline{49.8} & 3.9 & 6.0 & 7.2 \\
Qwen-2.5 7B Instruct & 31.5 & 51.6 & 63.6 & 9.3 & 14.2 & 19.3 & 0.6 & 0.7 & 1.1 \\
Qwen-2.5 14B Instruct & 64.0 & 78.0 & 83.9 & 15.4 & 21.1 & 26.3 & 0.5 & 1.3 & 1.9 \\
Qwen-2.5 32B Instruct & 67.9 & 83.6 & 88.4 & 20.8 & 32.9 & 42.5 & 2.9 & 3.5 & 4.0 \\
Qwen-2.5 72B Instruct & 75.4 & 82.2 & 88.4 & 17.4 & 26.4 & 35.6 & 0.2 & 2.2 & 2.8 \\
GPT-4o Mini & \underline{84.5} & \underline{93.6} & \underline{95.6} & 19.4 & 36.7 & 45.3 & \underline{6.5} & \underline{8.5} & \underline{10.7} \\
GPT-o3 Mini & \bfseries 99.6 & \bfseries 99.9 & \bfseries 99.9 & \bfseries 91.6 & \bfseries 96.8 & \bfseries 98.3 & \bfseries 37.4 & \bfseries 50.8 & \bfseries 57.5 \\
\bottomrule
\end{tabular}
}
\caption{3-Turns Accuracy (\%) of Anagram Scribble.}
\label{results_3turns_anagram}
\end{table*}

\begin{table*}[!ht]
\centering
\resizebox{.8\textwidth}{!}{
\begin{tabular}{lrrr|rrr|rrr}
\toprule
 & \multicolumn{9}{c}{\bfseries Password} \\
 & \multicolumn{3}{c|}{Easy} & \multicolumn{3}{c|}{Med} & \multicolumn{3}{c}{Hard} \\
\multicolumn{1}{l}{\textbf{Model} \hskip4em Turn \#} & \#1 & \#2 & \#3 & \#1 & \#2 & \#3 & \#1 & \#2 & \#3 \\
\midrule
Gemma-2 9B Instruct & 35.6 & 47.0 & 50.1 & 15.4 & 21.4 & 23.3 & 5.2 & 8.2 & 9.5 \\
Gemma-2 27B Instruct & 57.7 & 68.2 & 73.1 & 26.9 & 36.0 & 42.1 & 8.0 & 14.6 & 19.3 \\
Llama-3.1 8B Instruct & 29.1 & 42.6 & 50.5 & 9.6 & 15.2 & 19.0 & 2.5 & 4.5 & 5.5 \\
Llama-3.1 70B Instruct & 58.1 & 73.2 & 79.4 & 27.5 & 40.8 & \underline{47.7} & 8.3 & 15.7 & 21.5 \\
Llama-3.3 70B Instruct & \underline{60.1} & \underline{74.0} & \underline{81.0} & \underline{29.2} & \underline{40.9} & 47.4 & \underline{10.3} & \underline{16.9} & \underline{21.6} \\
Qwen-2.5 7B Instruct & 37.7 & 45.1 & 47.6 & 9.4 & 14.3 & 17.0 & 1.4 & 2.4 & 3.0 \\
Qwen-2.5 14B Instruct & 44.9 & 61.8 & 67.2 & 15.8 & 26.4 & 32.7 & 3.7 & 7.6 & 9.6 \\
Qwen-2.5 32B Instruct & 54.8 & 68.7 & 74.4 & 23.1 & 36.3 & 43.3 & 7.3 & 14.5 & 18.2 \\
Qwen-2.5 72B Instruct & 55.0 & 66.1 & 72.7 & 25.5 & 37.2 & 43.5 & 9.3 & 14.8 & 17.6 \\
GPT-4o Mini & 51.2 & 60.9 & 64.5 & 22.4 & 30.5 & 34.0 & 8.1 & 13.3 & 15.6 \\
GPT-o3 Mini & \bfseries 90.1 & \bfseries 96.2 & \bfseries 98.0 & \bfseries 74.6 & \bfseries 89.8 & \bfseries 91.7 & \bfseries 51.9 & \bfseries 70.9 & \bfseries 79.4 \\
\bottomrule
\end{tabular}
}
\caption{3-Turns Accuracy (\%) of Password Game.}
\label{results_3turns_password}
\end{table*}

\begin{table*}[!ht]
\centering
\resizebox{.8\textwidth}{!}{
\begin{tabular}{lrrr|rrr|rrr}
\toprule
 & \multicolumn{9}{c}{\bfseries Bracket} \\
 & \multicolumn{3}{c|}{Easy} & \multicolumn{3}{c|}{Med} & \multicolumn{3}{c}{Hard} \\
\multicolumn{1}{l}{\textbf{Model} \hskip4em Turn \#} & \#1 & \#2 & \#3 & \#1 & \#2 & \#3 & \#1 & \#2 & \#3 \\
\midrule
Gemma-2 9B Instruct & 26.6 & 57.7 & 67.1 & 3.5 & 9.3 & 12.4 & 0.0 & 0.0 & 0.0 \\
Gemma-2 27B Instruct & 27.4 & 87.5 & 92.9 & 11.6 & 40.2 & 56.4 & 0.1 & 0.2 & 0.4 \\
Llama-3.1 8B Instruct & 27.4 & 40.3 & 46.9 & 3.8 & 9.5 & 12.2 & 0.0 & 0.5 & 0.7 \\
Llama-3.1 70B Instruct & 42.7 & 81.6 & 94.3 & 23.9 & 43.7 & 68.9 & \underline{0.3} & 4.3 & \underline{19.2} \\
Llama-3.3 70B Instruct & 46.5 & 87.3 & \underline{96.3} & 20.4 & 48.9 & 72.7 & 0.0 & \underline{6.6} & 17.4 \\
Qwen-2.5 7B Instruct & 30.6 & 44.3 & 51.6 & 2.7 & 6.7 & 9.2 & 0.0 & 0.0 & 0.0 \\
Qwen-2.5 14B Instruct & 45.7 & 61.3 & 69.9 & 10.4 & 16.1 & 23.9 & 0.0 & 0.1 & 0.8 \\
Qwen-2.5 32B Instruct & \underline{66.5} & 82.1 & 87.1 & 25.1 & 41.4 & 50.1 & 0.1 & 2.8 & 6.2 \\
Qwen-2.5 72B Instruct & 65.9 & \underline{88.9} & 92.9 & \underline{39.3} & \underline{60.1} & \underline{74.7} & \underline{0.3} & 4.7 & 11.4 \\
GPT-4o Mini & 51.5 & 76.4 & 84.6 & 14.1 & 37.4 & 49.3 & 0.0 & 2.3 & 5.0 \\
GPT-o3 Mini & \bfseries 97.3 & \bfseries 99.8 & \bfseries 99.9 & \bfseries 84.9 & \bfseries 98.3 & \bfseries 99.6 & \bfseries 21.5 & \bfseries 63.6 & \bfseries 77.2 \\
\bottomrule
\end{tabular}
}
\caption{3-Turns Accuracy (\%) of Bracket Game.}
\label{results_3turns_bracket}
\end{table*}

\begin{table*}[!ht]
\centering
\resizebox{.8\textwidth}{!}{
\begin{tabular}{lrrr|rrr|rrr}
\toprule
 & \multicolumn{9}{c}{\bfseries String Search} \\
 & \multicolumn{3}{c|}{Easy} & \multicolumn{3}{c|}{Med} & \multicolumn{3}{c}{Hard} \\
\multicolumn{1}{l}{\textbf{Model} \hskip4em Turn \#} & \#1 & \#2 & \#3 & \#1 & \#2 & \#3 & \#1 & \#2 & \#3 \\
\midrule
Gemma-2 9B Instruct & 33.5 & 46.8 & 56.0 & 15.4 & 22.0 & 28.4 & 3.0 & 4.2 & 4.9 \\
Gemma-2 27B Instruct & 40.6 & 60.2 & 69.7 & 16.6 & 32.5 & 42.7 & 3.6 & 7.2 & 10.1 \\
Llama-3.1 8B Instruct & 33.5 & 46.5 & 54.2 & 8.1 & 11.1 & 12.2 & 1.4 & 1.8 & 1.9 \\
Llama-3.1 70B Instruct & 46.6 & 66.0 & 74.5 & 25.3 & 41.2 & 49.2 & 6.5 & 10.9 & 14.1 \\
Llama-3.3 70B Instruct & 44.0 & 64.9 & \underline{75.8} & 27.8 & 45.3 & \underline{56.4} & 6.2 & 11.3 & 15.0 \\
Qwen-2.5 7B Instruct & 22.7 & 37.5 & 47.2 & 12.1 & 21.6 & 28.5 & 3.0 & 4.0 & 5.0 \\
Qwen-2.5 14B Instruct & 28.8 & 48.5 & 59.0 & 13.3 & 23.7 & 31.7 & 2.6 & 3.5 & 4.5 \\
Qwen-2.5 32B Instruct & 44.4 & 61.4 & 70.0 & 25.7 & 41.0 & 50.4 & 5.8 & 10.0 & 12.8 \\
Qwen-2.5 72B Instruct & 46.4 & 64.3 & 71.8 & 25.4 & 37.7 & 43.3 & 5.9 & 9.6 & 12.6 \\
GPT-4o Mini & \underline{51.2} & \underline{66.2} & 71.4 & \underline{32.5} & \underline{45.9} & 52.6 & \underline{10.8} & \underline{13.4} & \underline{15.1} \\
GPT-o3 Mini & \bfseries 99.2 & \bfseries 99.8 & \bfseries 99.9 & \bfseries 97.9 & \bfseries 100.0 & \bfseries 100.0 & \bfseries 69.8 & \bfseries 92.8 & \bfseries 98.1 \\
\bottomrule
\end{tabular}
}
\caption{3-Turns Accuracy (\%) of String Search.}
\label{results_3turns_string}
\end{table*}

\begin{table*}[!ht]
\centering
\resizebox{.8\textwidth}{!}{
\begin{tabular}{lrrr|rrr|rrr}
\toprule
 & \multicolumn{9}{c}{\bfseries Crossword} \\
 & \multicolumn{3}{c|}{Easy} & \multicolumn{3}{c|}{Med} & \multicolumn{3}{c}{Hard} \\
\multicolumn{1}{l}{\textbf{Model} \hskip4em Turn \#} & \#1 & \#2 & \#3 & \#1 & \#2 & \#3 & \#1 & \#2 & \#3 \\
\midrule
Gemma-2 9B Instruct & 2.1 & 2.4 & 2.4 & 0.0 & 0.0 & 0.0 & 0.0 & 0.0 & 0.0 \\
Gemma-2 27B Instruct & 7.1 & 10.3 & 13.2 & 0.5 & 0.5 & 0.5 & 0.0 & 0.0 & 0.0 \\
Llama-3.1 8B Instruct & 2.2 & 3.7 & 5.3 & 0.0 & 0.0 & 0.0 & 0.0 & 0.0 & 0.0 \\
Llama-3.1 70B Instruct & 8.9 & 17.5 & \underline{26.0} & 0.4 & 0.5 & 0.8 & \underline{0.1} & \underline{0.1} & \underline{0.1} \\
Llama-3.3 70B Instruct & \underline{9.7} & \underline{17.7} & 25.6 & \underline{0.6} & 0.6 & \underline{1.0} & 0.0 & 0.0 & 0.0 \\
Qwen-2.5 7B Instruct & 2.2 & 3.1 & 3.2 & 0.2 & 0.2 & 0.2 & 0.0 & 0.0 & 0.0 \\
Qwen-2.5 14B Instruct & 0.0 & 0.6 & 1.4 & 0.0 & 0.0 & 0.0 & 0.0 & 0.0 & 0.0 \\
Qwen-2.5 32B Instruct & 2.3 & 4.2 & 5.1 & 0.0 & 0.0 & 0.0 & 0.0 & 0.0 & 0.0 \\
Qwen-2.5 72B Instruct & 5.2 & 10.4 & 14.6 & 0.0 & 0.0 & 0.0 & 0.0 & 0.0 & 0.0 \\
GPT-4o Mini & 6.9 & 10.6 & 12.9 & 0.4 & \underline{0.9} & \underline{1.0} & 0.0 & 0.0 & 0.0 \\
GPT-o3 Mini & \bfseries 57.9 & \bfseries 80.0 & \bfseries 95.4 & \bfseries 8.0 & \bfseries 28.2 & \bfseries 49.3 & \bfseries 0.7 & \bfseries 4.3 & \bfseries 11.8 \\
\bottomrule
\end{tabular}
}
\caption{3-Turns Accuracy (\%) of Crossword Arranger.}
\label{results_3turns_crossword}
\end{table*}

\begin{table*}[!ht]
\centering
\resizebox{.8\textwidth}{!}{
\begin{tabular}{lrrr|rrr|rrr}
\toprule
 & \multicolumn{9}{c}{\bfseries Sudoku} \\
 & \multicolumn{3}{c|}{Easy} & \multicolumn{3}{c|}{Med} & \multicolumn{3}{c}{Hard} \\
\multicolumn{1}{l}{\textbf{Model} \hskip4em Turn \#} & \#1 & \#2 & \#3 & \#1 & \#2 & \#3 & \#1 & \#2 & \#3 \\
\midrule
Gemma-2 9B Instruct & 25.9 & 30.0 & 31.7 & 4.8 & 6.8 & 8.3 & 0.0 & 0.0 & 0.0 \\
Gemma-2 27B Instruct & 38.8 & 46.3 & 48.9 & 13.6 & 17.7 & 19.1 & 0.0 & 0.0 & 0.0 \\
Llama-3.1 8B Instruct & 0.1 & 0.2 & 0.2 & 0.0 & 0.0 & 0.0 & 0.0 & 0.0 & 0.0 \\
Llama-3.1 70B Instruct & 14.7 & 24.8 & 29.1 & 2.8 & 5.0 & 6.9 & 0.0 & 0.0 & 0.0 \\
Llama-3.3 70B Instruct & 12.9 & 20.7 & 27.1 & 3.1 & 4.6 & 6.3 & 0.0 & 0.0 & 0.0 \\
Qwen-2.5 7B Instruct & 18.0 & 21.8 & 23.5 & 4.8 & 5.8 & 6.2 & 0.0 & 0.0 & 0.0 \\
Qwen-2.5 14B Instruct & 35.9 & 44.1 & 47.3 & 17.7 & 22.1 & 24.9 & 0.3 & 0.3 & 0.3 \\
Qwen-2.5 32B Instruct & 41.7 & 47.5 & 49.8 & 22.0 & 25.7 & 28.1 & \underline{0.4} & \underline{0.4} & \underline{0.4} \\
Qwen-2.5 72B Instruct & \underline{43.0} & \underline{49.5} & \underline{52.3} & \underline{22.1} & \underline{28.0} & \underline{29.9} & \underline{0.4} & \underline{0.4} & \underline{0.4} \\
GPT-4o Mini & 25.5 & 31.6 & 34.7 & 4.9 & 7.9 & 9.5 & 0.0 & 0.0 & 0.0 \\
GPT-o3 Mini & \bfseries 99.2 & \bfseries 99.7 & \bfseries 100.0 & \bfseries 80.9 & \bfseries 99.7 & \bfseries 100.0 & \bfseries 2.0 & \bfseries 3.4 & \bfseries 5.5 \\
\bottomrule
\end{tabular}
}
\caption{3-Turns Accuracy (\%) of Text Sudoku.}
\label{results_3turns_sudoku}
\end{table*}

\begin{table*}[!ht]
\centering
\resizebox{.8\textwidth}{!}{
\begin{tabular}{lrrr|rrr|rrr}
\toprule
 & \multicolumn{9}{c}{\bfseries Islands} \\
 & \multicolumn{3}{c|}{Easy} & \multicolumn{3}{c|}{Med} & \multicolumn{3}{c}{Hard} \\
\multicolumn{1}{l}{\textbf{Model} \hskip4em Turn \#} & \#1 & \#2 & \#3 & \#1 & \#2 & \#3 & \#1 & \#2 & \#3 \\
\midrule
Gemma-2 9B Instruct & 22.8 & 30.2 & 31.8 & 2.9 & 3.1 & 3.6 & 0.3 & 0.4 & 0.9 \\
Gemma-2 27B Instruct & 14.5 & 22.5 & 31.4 & 5.8 & 7.4 & 8.7 & 1.9 & 2.9 & 3.5 \\
Llama-3.1 8B Instruct & 3.5 & 5.9 & 6.4 & 2.7 & 3.3 & 3.4 & 0.1 & 0.1 & 0.1 \\
Llama-3.1 70B Instruct & 31.2 & 35.4 & 36.4 & 9.1 & 14.7 & 17.0 & 1.0 & 3.5 & 5.2 \\
Llama-3.3 70B Instruct & \underline{45.8} & \underline{58.8} & \underline{63.1} & 8.1 & 13.6 & 18.1 & \underline{3.1} & \underline{5.7} & \underline{7.7} \\
Qwen-2.5 7B Instruct & 6.9 & 7.3 & 8.2 & 2.4 & 3.9 & 3.9 & 0.3 & 0.4 & 0.6 \\
Qwen-2.5 14B Instruct & 8.2 & 12.1 & 14.7 & 1.2 & 1.6 & 2.0 & 0.5 & 1.2 & 2.0 \\
Qwen-2.5 32B Instruct & 43.3 & 56.4 & 60.6 & 8.1 & 11.4 & 16.1 & 2.2 & 3.3 & 4.9 \\
Qwen-2.5 72B Instruct & 35.8 & 48.0 & 56.7 & \underline{13.0} & \underline{18.8} & \underline{21.2} & 2.5 & 5.1 & 7.1 \\
GPT-4o Mini & 19.6 & 31.0 & 32.3 & 9.1 & 11.9 & 13.8 & 1.9 & 3.7 & 4.5 \\
GPT-o3 Mini & \bfseries 95.5 & \bfseries 100.0 & \bfseries 100.0 & \bfseries 81.1 & \bfseries 95.9 & \bfseries 98.9 & \bfseries 57.4 & \bfseries 80.5 & \bfseries 88.1 \\
\bottomrule
\end{tabular}
}
\caption{3-Turns Accuracy (\%) of Islands.}
\label{results_3turns_islands}
\end{table*}

\begin{table*}[!ht]
\centering
\resizebox{.8\textwidth}{!}{
\begin{tabular}{lrrr|rrr|rrr}
\toprule
 & \multicolumn{9}{c}{\bfseries Ordering} \\
 & \multicolumn{3}{c|}{Easy} & \multicolumn{3}{c|}{Med} & \multicolumn{3}{c}{Hard} \\
\multicolumn{1}{l}{\textbf{Model} \hskip4em Turn \#} & \#1 & \#2 & \#3 & \#1 & \#2 & \#3 & \#1 & \#2 & \#3 \\
\midrule
Gemma-2 9B Instruct & 25.5 & 59.0 & 81.0 & 4.0 & 10.3 & 19.8 & 0.9 & 2.6 & 5.6 \\
Gemma-2 27B Instruct & 25.4 & 64.7 & 83.9 & 3.8 & 11.6 & 25.5 & \underline{1.5} & 3.3 & \underline{6.9} \\
Llama-3.1 8B Instruct & 20.4 & 41.0 & 71.0 & 2.8 & 4.8 & 11.5 & 0.5 & 1.2 & 2.8 \\
Llama-3.1 70B Instruct & 24.5 & 56.4 & 79.1 & 3.6 & 12.3 & 20.9 & 1.0 & 3.4 & 6.6 \\
Llama-3.3 70B Instruct & 24.3 & 54.8 & 77.8 & 4.2 & 10.4 & 17.3 & 0.7 & 3.1 & 5.7 \\
Qwen-2.5 7B Instruct & 22.8 & 50.6 & 70.4 & 3.0 & 9.2 & 15.8 & 0.9 & 2.5 & 4.5 \\
Qwen-2.5 14B Instruct & 27.4 & 63.7 & 82.3 & 4.1 & 14.1 & 23.1 & 0.5 & 2.8 & 4.8 \\
Qwen-2.5 32B Instruct & \underline{31.9} & \underline{69.3} & \underline{84.0} & 5.7 & \underline{18.5} & \underline{27.6} & 0.4 & 3.1 & 6.8 \\
Qwen-2.5 72B Instruct & 28.8 & 64.7 & 80.8 & \underline{6.5} & 16.0 & 25.5 & 0.8 & 2.8 & 6.0 \\
GPT-4o Mini & 30.6 & 65.6 & 83.8 & 5.8 & 17.3 & 25.4 & 0.9 & \underline{3.7} & 5.9 \\
GPT-o3 Mini & \bfseries 98.6 & \bfseries 99.9 & \bfseries 100.0 & \bfseries 96.0 & \bfseries 99.0 & \bfseries 99.5 & \bfseries 22.5 & \bfseries 73.8 & \bfseries 89.0 \\
\bottomrule
\end{tabular}
}
\caption{3-Turns Accuracy (\%) of Ordering Text.}
\label{results_3turns_ordering}
\end{table*}

\clearpage


\section{Prompt Templates and Games Constraints}
\label{sec:task_details}

We detail the prompt templates and constraints for prompt constructions here: Anagram Scribble in Table~\ref{tab:anagram}, Password Games in Table~\ref{tab:password}, Bracket Game in Table~\ref{tab:bracket}, String Search in Table~\ref{tab:string}, Crossword Arranger in Table~\ref{tab:crossword}, Text Sudoku in Table~\ref{tab:sudoku}, Islands in Table~\ref{tab:islands}, and Ordering Text in Table~\ref{tab:ordering}.

\begin{table*}[!th]
\centering\small
\resizebox{\textwidth}{!}{
    \begin{tabular}{p{15cm}}
    \toprule
    \textbf{<Prompt Template ($\mathcal{P}$)>} \\
    \\
    \texttt{Construct a valid \textbf{[N]}-character English word from the following letters:}\\
    \texttt{`\textbf{[C\textsubscript{1}]}', `\textbf{[C\textsubscript{2}]}', $\ldots$, `\textbf{[C\textsubscript{N+M}]}'.}\\
    \texttt{Each character can be used multiple times. Please write None if there is no valid combination. Print only the answer.}\\
    \midrule
    \textbf{<Example>} \\
    \\
    \emph{Constraints ($\mathcal{C}$)}:\\
    \texttt{ - \textbf{[N]=6}-character English word.}\\
    \texttt{ - Letters \textbf{[C\textsubscript{1$\ldots$8}]} = `e', `l', `o', `d', `p', `h', `i'.}\\
    \\
    \emph{Possible Answer}:\\
    \texttt{hoodie}\\
    \bottomrule
    \end{tabular}
    
}
\caption{Anagram Scribble.}
\label{tab:anagram}
\end{table*}

\begin{table*}
\centering\small
\resizebox{\textwidth}{!}{
    \begin{tabular}{p{\textwidth}}
    \toprule
    \textbf{<Prompt Template ($\mathcal{P}$)>}\\
    \\
    \texttt{Please write a text string without any space by following a set of given rules. Please write only the answer and follow the following criteria:}\\
    \texttt{ - the text has \textbf{[C\textsubscript{1}]}}\\
    \texttt{  $\cdots$}\\
    \texttt{ - the text has \textbf{[C\textsubscript{$\alpha$}]}}\\
    
    \midrule
    \textbf{<Example>} \\
    \\
    \emph{Constraints ($\mathcal{C}$)}:\\
    \texttt{ - \textbf{[C\textsubscript{1}]} = 6 English characters}\\
    \texttt{ - \textbf{[C\textsubscript{2}]} = 0 uppercase character}\\
    \\
    \emph{Possible Answer}:\\
    \texttt{hoodie}\\
    \end{tabular}
}
\resizebox{\textwidth}{!}{
    \begin{tabular}{p{.78\textwidth}cc}
    \midrule
    \textbf{<Possible Rules [C\textsubscript{$\chi$}]>} & \textbf{<Type>} & \textbf{<Repeatable>}\\
        \texttt{ - only \textbf{[N]} characters} & counting & no \\ 
        \texttt{ - \textbf{[N]} uppercase characters} & counting & no \\
        \texttt{ - \textbf{[N]} lowercase characters} & counting & no \\
        \texttt{ - \textbf{[N]} latin character} & counting & no \\
        \texttt{ - \textbf{[N]} number digits} & counting & no \\
        \texttt{ - \textbf{[N]} number of roman digits} & counting & no \\
        \texttt{ - \textbf{[N]} special characters, including '!', '@', '\#', '\$', '\%', '\^', '\&', '*'} & counting & no \\ 
        \texttt{ - \textbf{[N]} \textbf{[Ch]} character} & counting & yes \\ 
        \texttt{ - \textbf{[S]} string} & string-matching & yes \\
        \texttt{ - the capital city of \textbf{[S]}} & knowledge & yes \\
        \texttt{ - the continent of \textbf{[S]}} & knowledge & yes \\
        \texttt{ - a number that equals to \textbf{[E\textsubscript{\text{math}}]}} & math & yes \\
        \texttt{ - a number that equals to \textbf{[E\textsubscript{\text{word}}]}} & math & yes \\
    \midrule
        \textbf{<Parameters>} \\
        \texttt{ $\cdot$ \textbf{[N]}} $\in$ $\mathbb{Z^+}$; \quad
        \texttt{\textbf{[Ch]}} $\in$ \{`A'$\ldots$`Z', `a'$\ldots$`z'\}; &&\\
        \texttt{ $\cdot$ \textbf{[S]}} is a random English word; &&\\
        \texttt{ $\cdot$ \textbf{[E\textsubscript{\text{math}}]}} is an arithmetical expression written in number and symbols, e.g. ``\texttt{4 + 2}''; &&\\
        \texttt{ $\cdot$ \textbf{[E\textsubscript{\text{word}}]}} is an arithmetical expression written in words, e.g. ``\texttt{four plus two}''; &&\\
    \bottomrule
    \end{tabular}
}
\caption{Password Game.}
\label{tab:password}
\end{table*}

\begin{table*}[!th]
    \centering\small
    \begin{tabular}{p{15cm}}
    \toprule
    \textbf{<Prompt Template ($\mathcal{P}$)>}\\
    \\
    \texttt{You are given a text \textbf{[S]} Your job is to put some valid parenthesis brackets in the text such that:}\\
    \texttt{ - \textbf{[W\textsubscript{1}]} is inside a \textbf{[B\textsubscript{1}]} bracket}\\
    \texttt{  $\cdots$}\\
    \texttt{ - \textbf{[W\textsubscript{N}]} is inside a \textbf{[B\textsubscript{N}]} bracket}\\ \\
    \texttt{The open and close parenthesis for block is [ ], curly is { }, round is ( ), and angle is < >.}\\
    \texttt{The bracket depth must be \textbf{[D]} and print only the answer}\\

    \midrule
    \textbf{<Example>}\\
    \\
    \emph{Constraints ($\mathcal{C}$)}:\\
    \texttt{The text is \textbf{[S] = `fabuloustextgames'}, and \textbf{[W] = [`games', `text', `fabulous']} are inside \textbf{[B] = [round, angle, block]} bracket, respectively. Depth must be \textbf{[D] = 2}.}\\
    \\

    \emph{Possible Answer}:\\
    \texttt{\{[fabulous]<text>(games)\}}\\
    
    \bottomrule
    \end{tabular}
    \caption{Bracket Game}
    \label{tab:bracket}
\end{table*}

\begin{table*}
\centering\small
\resizebox{\textwidth}{!}{
    \begin{tabular}{p{15cm}}
    \toprule
    \textbf{<Prompt Template ($\mathcal{P}$)>} \\
    \\
    \texttt{You are given the following string:}\\
    \texttt{\textbf{[S]}}\\ \\
    \texttt{Find a substring of exactly \textbf{[N]} characters long that:}\\
    \texttt{ - Contains \textbf{[X\textsubscript{1}$\ldots$\textsubscript{$\alpha$}]}}\\
    \texttt{ - Does not contain \textbf{[Y\textsubscript{1}$\ldots$\textsubscript{$\beta$}]}}\\
    \texttt{ - \textbf{[Z\textsubscript{1}]}}\\
    \texttt{ $\cdots$}\\
    \texttt{ - \textbf{[Z\textsubscript{$\gamma$}]}}\\ \\
    \texttt{Print only the answer.}\\
    
    \midrule
    \textbf{<Example>} \\
    \\
    \emph{Constraints ($\mathcal{C}$)}:\\
    \texttt{ - \textbf{[S]} = ``hengooserabbitant''}\\
    \texttt{ - \textbf{[X\textsubscript{1$\ldots$1}]} = \{`g'\}}\\
    \texttt{ - \textbf{[Y\textsubscript{1$\ldots$2}]} = \{`i', `a'\}}\\
    \texttt{ - No complex rules \textbf{[Z]} = $\emptyset$}\\
    \\
    \emph{Possible Answer}:\\
    \texttt{goo}\\
    
    \end{tabular}
}
\resizebox{\textwidth}{!}{
    \begin{tabular}{p{.6\textwidth}c}
    \midrule
    \textbf{<Possible Complex Rules [Z\textsubscript{$\chi$}]>} & \textbf{Mutually Exclusive Group} \\
    \texttt{ - forms a palindrome} & - \\
    \texttt{ - has 2 consecutive consonants} & $\alpha$ \\
    \texttt{ - does not have 2 consecutive consonants} & $\alpha$ \\
    \texttt{ - has 2 consecutive vowels} & $\beta$ \\
    \texttt{ - does not have 2 consecutive vowels} & $\beta$ \\
    \texttt{ - has more vowels than consonants} & $\gamma$ \\
    \texttt{ - has less vowels than consonants} & $\gamma$ \\
    \texttt{ - has the same amount of vowels and consonants} & $\gamma$ \\
    \bottomrule
    \end{tabular}
}
\caption{String Search.}
\label{tab:string}
\end{table*}

\begin{table*}
\centering\small
\resizebox{\textwidth}{!}{
    \begin{tabular}{p{15cm}}
    \toprule
    \textbf{<Prompt Template ($\mathcal{P}$)>} \\
    \\
    \texttt{Given a board size of \textbf{[N]}x\textbf{[N]}, arrange a possible crossword puzzle answer from a list of words.}
    \texttt{Item in the list can only be used once.}\\ \\
    \texttt{List of words:}\\
    \texttt{ - \textbf{[W\textsubscript{1}]}}\\
    \texttt{ - \textbf{[W\textsubscript{2}]}}\\
    \texttt{  $\cdots$}\\ \\
    \texttt{Print only the answer.}
    \\
    \midrule
    \textbf{<Example>} \\
    \\
    \emph{Constraints ($\mathcal{C}$)}:\\
    \texttt{ - \textbf{[N] = 3}  (3x3 grid)}\\
    \texttt{ - \textbf{[W\textsubscript{1$\ldots$8}]} = \{app, all, and, lee, let, pat, pee, pet\}}\\
    \\
    \emph{Possible Answer}:\\
    \texttt{app}\\
    \texttt{lee}\\
    \texttt{let}\\
    \bottomrule
    \end{tabular}
}
\caption{Crossword Arranger.}
\label{tab:crossword}
\end{table*}

\begin{table*}
\centering\small
\resizebox{\textwidth}{!}{
    \begin{tabular}{p{15cm}}
    \toprule
    \textbf{<Prompt Template ($\mathcal{P}$)>} \\
    \\
    \texttt{Please solve the \textbf{[N]}x\textbf{[N]} sudoku with \textbf{[V]} as the values and fill \_ with the possible value and only print the answer. Follow the sudoku rule.}\\
    \texttt{\textbf{[S\textsubscript{1,1}]}$\ldots$\textbf{[S\textsubscript{1,N}]} $\cdots$ \textbf{[S\textsubscript{N,1}]}$\ldots$\textbf{[S\textsubscript{N,N}]}}\\
    \midrule
    \textbf{<Example>} \\
    \\
    \emph{Constraints ($\mathcal{C}$)}:\\
    \texttt{ - \textbf{[N] = 4}  (4x4 grid)}\\
    \texttt{ - \textbf{[V]} = \{A, B, C, D\}}\\
    \texttt{ - \texttt{\textbf{[S\textsubscript{1,1}]}$\ldots$\textbf{[S\textsubscript{N,N}]}} = ``A\_CD CD\_B \_AD\_ DCBA''}\\
    \\
    \emph{Possible Answer}:\\
    \texttt{ABCD}\\
    \texttt{CDAB}\\
    \texttt{BADC}\\
    \texttt{DCBA}\\
    \bottomrule
    \end{tabular}
}
\caption{Text Sudoku.}
\label{tab:sudoku}
\end{table*}


\begin{table*}[!th]
\centering\small
    \begin{tabular}{p{15cm}}
    \toprule
    \textbf{<Prompt Template ($\mathcal{P}$)>} \\
    \\
    \texttt{You are asked to construct a 2D \textbf{[N]} x \textbf{[N]} grid, consisting of water tiles (denoted by '.'), land tiles (denoted by '\#'), and coconut tree tiles (denoted by 'o'). 
    Coconut tree tiles are also considered as land tiles.} \\ \\
    \texttt{A group of connected land tiles in 4 cardinal directions forms an island.} \\ \\
    \texttt{Your 2D grid must follow the following rules:} \\
    \texttt{ - There must be exactly \textbf{[K]} islands.} \\
    \texttt{ - The size of each island must be from \textbf{[Y\textsubscript{min}]} to \textbf{[Y\textsubscript{max}]} tiles each.} \\
    \texttt{ - There must be exactly \textbf{[L]} islands that have coconut trees on them.} \\
    \texttt{ - There must be exactly \textbf{[C]} total coconut trees.} \\
    \\
    \texttt{Print only the answer.}\\
    
    
    \midrule
    \textbf{<Example>} \\
    \\
    \emph{Constraints ($\mathcal{C}$)}:\\
    \texttt{ - \textbf{[N] = 6} (6x6 grid),}\\
    \texttt{ - \textbf{[K] = 3} islands,}\\
    \texttt{ - island size from \textbf{[Y\textsubscript{min}]=5} to \textbf{[Y\textsubscript{max}]=10} tiles,}\\
    \texttt{ - \textbf{[L] = 2} islands hhave coconut trees,} \\
    \texttt{ - \textbf{[C] = 4} coconut trees in total.}\\
    \\
    \emph{Possible Answer}:\\
    \texttt{.\#\#...}\\
    \texttt{\#o\#...}\\
    \texttt{.o\#.\#\#}\\
    \texttt{....\#\#}\\
    \texttt{\#o\#..\#}\\
    \texttt{\#o\#\#..}\\

    
    \bottomrule
    \end{tabular}
\caption{Islands.}
\label{tab:islands}
\end{table*}

\begin{table*}[!th]
\centering\small
\resizebox{\textwidth}{!}{
    \begin{tabular}{p{15cm}}
    \toprule
    \textbf{<Prompt Template ($\mathcal{P}$)>} \\
    \\
    \texttt{Given a set of rules to calculate point, sort the set of words in decreasing order.}\\
    \texttt{When there 2 or more words with same point, sort lexicographically.}\\
    \texttt{Rules:}\\
    \texttt{ - \textbf{[C\textsubscript{1}]}} gets \textbf{[P\textsubscript{1}]} points\\
    \texttt{ - add \textbf{[P\textsubscript{2}]} points if \textbf{[C\textsubscript{2}]}}\\
    \texttt{  $\cdots$}\\
    \texttt{Words:}\\
    \texttt{ - \textbf{[W\textsubscript{1}]}}\\
    \texttt{ - \textbf{[W\textsubscript{2}]}}\\
    \texttt{  $\cdots$}\\
    \texttt{Print only the answer.}\\
    

    \midrule
    \textbf{<Example>} \\
    \\
    \emph{Constraints ($\mathcal{C}$)}:\\
    \texttt{Rules:}\\
    \texttt{ - add \textbf{[P\textsubscript{1}]=1} point if \textbf{[C\textsubscript{1}] = there exists 'g' in the word}}\\
    \texttt{ - \textbf{[C\textsubscript{1}] = word less than 5 characters} gets \textbf{[P\textsubscript{1}]=10} points}\\
    \texttt{Words:}\\
    \texttt{ - \textbf{[W\textsubscript{1}]} = hen}\\
    \texttt{ - \textbf{[W\textsubscript{2}]} = goose}\\
    \texttt{ - \textbf{[W\textsubscript{3}]} = rabbit}\\
    \texttt{ - \textbf{[W\textsubscript{4}]} = ant}\\
    \\
    \emph{Possible Answer}:\\
    \texttt{ant}\\
    \texttt{hen}\\
    \texttt{goose}\\
    \texttt{rabbit}\\

        
    \end{tabular}
}
\resizebox{\textwidth}{!}{
    \begin{tabular}{p{.85\textwidth}c}
    \midrule
    \textbf{<Possible Rules Condition [Z\textsubscript{$\chi$}]>} & \textbf{<Type>} \\
    \texttt{ - every (vowel | consonant)} & Counting \\
    \texttt{ - every vowel right after a consonant} & Pattern \\
    \texttt{ - every consonant right after a vowel} & Pattern \\
    \texttt{ - every pair of consecutive vowels} & Pattern \\
    \texttt{ - every \textbf{[N]} consecutive vowels} & Pattern \\
    \texttt{ - every pair of consecutive consonants} & Pattern \\
    \texttt{ - every \textbf{[N]} consecutive consonants} & Pattern \\
    
    \texttt{ - word that has exactly \textbf{[N]} characters} & Counting \\
    \texttt{ - word not equal to \textbf{[N]} characters} & Counting \\
    \texttt{ - word (more | less) than \textbf{[N]} characters} & Counting \\
    \texttt{ - word more than \textbf{[N$_1$]} characters and less than \textbf{[N$_2$]} characters} & Counting \\
    \texttt{ - word (more | less) than \textbf{[N$_1$]} characters but not equal to \textbf{[N$_2$]} characters} & Counting \\
    
    \texttt{ - word (starts | ends) with `\textbf{[S]}'} & String-search \\
    \texttt{ - word starts with `\textbf{[S$_1$]}' and ends with `\textbf{[S$_1$]}'} & String-search \\
    \texttt{ - there exists `\textbf{[S]}' in the word} & String-search \\
    \texttt{ - there exists exactly \textbf{[N]} `\textbf{[S]}' in the word} & String-search \\

    \midrule
        \textbf{<Parameters>} \\
        \texttt{ $\cdot$ \textbf{[N]}} $\in$ $\mathbb{Z^+}$; &\\
        \texttt{ $\cdot$ \textbf{[S]}} is any character sequence; &\\

    \bottomrule
    \end{tabular}
}
\caption{Ordering Text}
\label{tab:ordering}
\end{table*}

\clearpage

\section{Longer Reasoning Leads to Confusion}
\label{sec:longer_reasoning_confusion}

\begin{table*}[!th]
\begin{tabular}{p{155mm}}
    \toprule\vspace{-1.em}
    \begin{lstlisting}[emphstyle={\bfseries\itshape\textcolor{red}},emph={can,only,be,used,once}]
        <think>
        Okay, so I need to create a 3x3 crossword puzzle using the given list of words: age, ago, dub, man, map, nor, per, pop. Each word can only be used once. Let me think about how to approach this.
    \end{lstlisting}
    \begin{lstlisting}[emphstyle={\bfseries\itshape\textcolor{red}},emph={each,word,will,occupy}]
        First, I should list out the words and their lengths. All of them are three letters, which fits perfectly into a 3x3 grid. That's good because each word will occupy a row or a column.
    \end{lstlisting}
    \begin{lstlisting}[emphstyle={\bfseries\itshape\textcolor{red}},emph={}]
        I should consider how these words can intersect. Since it's a crossword, the words will share letters where they cross. Let me look for words that share common letters, especially in the middle, because that's where they'll intersect in a 3x3 grid.
        
        Looking at the words:
        - age
        - ago
        - dub
        - man
        - map
        - nor
        - per
        - pop
    \end{lstlisting}
    \qquad$\cdots$\\\vspace{-.5em}
    \begin{lstlisting}
        Wait, let's try to place "nor" somewhere. "Nor" is N, O, R. Maybe place it vertically in column 3: N, O, R.

        So column 3: N, O, R.
        
        Then, row 1: _, _, N
        Row 2: _, _, O
        Row 3: _, _, R
        
        Now, let's see. Maybe "age" is vertical in column 2: A, G, E.
        
        So column 2: A, G, E.
        
        Then, row 1: _, A, N
        Row 2: _, G, O
        Row 3: _, E, R
        
        Now, let's see what's left: ago, dub, map, man, per, pop.
        
        Looking at row 1: _, A, N. Maybe "man" can go here: M, A, N. So row 1: M, A, N.
        
        Then, column 1: M, _, _. Maybe "map" vertically: M, A, P. But row 1, column 1 is M, so row 2, column 1 would be A, and row 3, column 1 would be P. So row 2: A, G, O; row 3: P, E, R.
        
        Now, let's check the words used: "nor" in column 3, "age" in column 2, "man" in row 1, "map" in column 1. Remaining words: ago, dub, per, pop.
        
        Looking at row 2: A, G, O. That's "ago", which is in our list. So row 2: A, G, O.
        
        Now, row 3: P, E, R. That's "per", which is in our list.
        
        So the grid would look like:
        
        Row 1: M, A, N
        Row 2: A, G, O
        Row 3: P, E, R
        
        But wait, "map" is M, A, P, but in this configuration, column 1 is M, A, P, which is "map". So that works.
        
        Now, let's check all words used: man, ago, map, per, nor. Wait, "nor" is N, O, R, but in this grid, column 3 is N, O, R, which is "nor". So that's correct.
        
        But we still have "dub" and "pop" left. Hmm, that's a problem because we've used all the words except those two, but they don't fit into the grid.
    \end{lstlisting}
    \begin{lstlisting}[emphstyle={\bfseries\itshape\textcolor{red}},emph={doesn't,use,all,which,is,required},alsoletter={'}]
        Wait, no, let's count: we have 8 words, and the grid uses 5 words: man, ago, map, per, nor. So we're missing "dub" and "pop". That means this configuration doesn't use all the words, which is required.

    \end{lstlisting}

    \qquad$\cdots$\\
    \bottomrule
\end{tabular}
\caption{DeepSeek-R1-Distill 14B Hallucinated despite getting the correct answer along the way and ended up changing the answer to the wrong one.}
\label{tab:deepseek_hallucinated}
\end{table*}

\section{Feedback Templates}
\label{sec:feedbacks}

\begin{table*}[!ht]
    \centering
    \resizebox{0.93\textwidth}{!}{
    \begin{tabular}{cp{155mm}}
        \toprule
        \multicolumn{1}{c}{\textbf{Game}} & \multicolumn{1}{c}{\textbf{Feedback}} \\
        
        \midrule
        \multirow[l]{4}{*}{\makecell{Anagram\\Scribble}}
        & Your answer must be exactly \texttt{<int>} characters long \\
        & Your answer must only contain the characters provided \\
        & Your answer must not contain repeated characters \\
        & Your answer is not a valid English word \\
        
        \midrule
        \multirow[l]{1}{*}{Password}
        & \texttt{<str\textsubscript{prediction}>} is not satisfying this rule: \texttt{<str\textsubscript{rule}>}. \\
        
        \midrule
        \multirow[l]{5}{*}{\makecell{Bracket\\Game}}
        & You are not allowed to change the character sequence of base text \texttt{<string>} \\
        & There is a closing bracket without an open bracket \\
        & The depth of the bracket is \texttt{<int>}. The expected depth is \texttt{<integer>} \\
        & The text '\texttt{<str\textsubscript{rule\_text}>}' is not found in your answer. \\
        & \makecell[l]{The text '\texttt{<str\textsubscript{rule\_text}>}' is not inside any \texttt{<`block'|`curly'|`round'|`angle'>} \\ \qquad bracket \texttt{<char\textsubscript{\text{bracket\_open}}>} \texttt{<char\textsubscript{\text{bracket\_close}}>}} \\
        
        \midrule
        \multirow[l]{12}{*}{\makecell{String\\Search}}
        & \texttt{<str\textsubscript{answer}>} is not \texttt{<int>} characters long. \\
        & \texttt{<str\textsubscript{answer}>} does not exist in \texttt{<str\textsubscript{input\_text}>}. \\
        & \texttt{<str\textsubscript{answer}>} does not have 2 consecutive consonants \\
        & \texttt{<str\textsubscript{answer}>} has 2 consecutive consonants \\
        & \texttt{<str\textsubscript{answer}>} does not have 2 consecutive vowels \\
        & \texttt{<str\textsubscript{answer}>} has 2 consecutive vowels \\
        & \texttt{<str\textsubscript{answer}>} has less or equal vowels than consonants \\
        & \texttt{<str\textsubscript{answer}>} has more or equal vowels than consonants \\
        & \texttt{<str\textsubscript{answer}>} does not have the same amount of vowels and consonants \\
        & \texttt{<char>} does not appear in \texttt{<str\textsubscript{answer}>}. \\
        & \texttt{<char>} exists in \texttt{<str\textsubscript{answer}>}. \\
        & \texttt{<str\textsubscript{answer}>} is not a palindrome. \\

        \midrule
        \multirow[l]{2}{*}{\makecell{Crossword\\Arranger}}
        & Mismatch answer length found!! Expected size of \texttt{<integer>}, got \texttt{<integer>}.\\
        & Mismatch answer word found!! \texttt{<`Horizontal'|`Vertical'>} word \texttt{<string>} is not in the word set. \\

        \midrule
        \multirow[l]{4}{*}{\makecell{Text\\Sudoku}}
        & There are unfilled cells \\
        & Your answer is wrong in shape, it should be \texttt{<int>}x\texttt{<int>} sudoku. \\
        & There are unrecognized characters, or possibly unfilled cells. \\
        & "One or more characters are replaced" \\
        
        \midrule
        \multirow[l]{6}{*}{Islands}
        & 2D grid is not \texttt{<int>} x \texttt{<int>}. (\texttt{<int\textsubscript{pred}>} x \texttt{<int\textsubscript{pred}>})\\
        & 2D contains invalid character (\texttt{<char>}) \\
        & There must be exactly \texttt{<int>} islands, but you provided \texttt{<int>} islands \\
        & The size of each island must be from \texttt{<int>} to \texttt{<int>} tiles \\
        & There must be exactly \texttt{<int>} islands that have coconut trees on them \\
        & There must be exactly \texttt{<int>} total coconut trees. \\
        
        \midrule
        \multirow[l]{2}{*}{\makecell{Ordering\\Text}}
        & Your answer is too short. There should be \texttt{<int>} items. \\
        & \texttt{<str\textsubscript{answer}>} is not supposed to be the \texttt{<str\textsubscript{ordinal\_number}>} word in the order. \\
        
        \bottomrule
    \end{tabular}
    }
    \caption{List of Feedback.}
    \label{tab:list-feedback}
\end{table*}

\end{document}